%% file: 00_main_AIAA.tex
\documentclass[journal]{new-aiaa}
\usepackage[utf8]{inputenc}

\usepackage[colorinlistoftodos]{todonotes}

\usepackage{graphicx} 

\usepackage{amsmath,amssymb,amsfonts}
\usepackage{mathtools}
\usepackage{bm}
\DeclareMathAlphabet{\mathpzc}{OT1}{pzc}{m}{it}

\input{src/newcommands}
\input{src/preamble}

\usepackage{algorithmic}
\usepackage{graphicx}
\usepackage{textcomp}
\usepackage{subcaption}
\usepackage{xcolor}
\usepackage[nolist]{acronym}
\usepackage{comment}

\usepackage{tikz}
\usetikzlibrary{positioning,fit}
\usetikzlibrary{calc}
\usetikzlibrary{arrows.meta}
\usetikzlibrary{tikzmark} 
\usetikzlibrary{shapes, shadows}
\usepackage{pgfplots}
\usepackage{siunitx}
\usepackage{longtable,tabularx}
\usepackage{tabularray}
\UseTblrLibrary{booktabs}

\DeclareSIUnit{\rad}{rad}
\DeclareSIUnit{\nanotesla}{nT}
\DeclareSIUnit{\hertz}{Hz}

\usepackage[version=4]{mhchem}
\pgfplotsset{compat=1.18} 
\usepackage{hyperref}

\title{Airborne Magnetic Anomaly Navigation with Neural-Network-Augmented Online Calibration}

\author{Antonia Hager\footnote{PhD Candidate, Airbus Central Research and Technology / Department of Engineering Cybernetics, Norwegian University of Science and Technology (NTNU), Antonia.Hager@airbus.com}, Sven Nebendahl\footnote{MS, Airbus Central Research and Technology, Taufkirchen, Germany}, Alexej Klushyn\footnote{PhD, Airbus Central Research and Technology, Taufkirchen, Germany.} and Jasper Krauser\footnote{PhD, Airbus Central Research and Technology, Germany.}}
\affil{Airbus, 82024 Taufkirchen, Germany.}
\author{Torleiv H. Bryne\footnote{Associate Professor, Department of Engineering Cybernetics (NTNU).}, Tor Arne Johansen\footnote {Professor, Department of Engineering Cybernetics (NTNU).}}
\affil{Norwegian University of Science and Technology, 7034 Trondheim, Norway.}

\hypersetup{pdftitle=Airborne Magnetic Anomaly Navigation with Neural-Network-Augmented Online Calibration, pdfauthor={Antonia Hager, Sven Nebendahl, Alexej Klushyn, Jasper Krauser, Torleiv H. Bryne, Tor Arne Johansen}}

\begin{document}
\input{src/acronyms}

\maketitle
\begin{abstract}
Airborne Magnetic Anomaly Navigation (MagNav) provides a jamming-resistant and robust alternative to satellite navigation but requires the real-time compensation of the aircraft platform's large and dynamic magnetic interference. State-of-the-art solutions often rely on extensive offline calibration flights or pre-training, creating a logistical barrier to operational deployment. We present a fully adaptive MagNav architecture featuring a ``cold-start'' capability that identifies and compensates for the aircraft’s magnetic signature entirely in-flight. The proposed method utilizes an extended Kalman filter with an augmented state vector that simultaneously estimates the aircraft’s kinematic states as well as the coefficients of the physics-based Tolles-Lawson calibration model and the parameters of a Neural Network to model aircraft interferences.
The Kalman filter update is mathematically equivalent to an online Natural Gradient descent, integrating superior convergence and data efficiency of state-of-the-art second-order optimization directly into the navigation filter. 
To enhance operational robustness, the neural network is constrained to a residual learning role, modeling only the nonlinearities uncorrected by the explainable physics-based calibration baseline. 
Validated on the MagNav Challenge dataset, our framework effectively bounds inertial drift using a magnetometer-only feature set. The results demonstrate navigation accuracy comparable to state-of-the-art models trained offline, without requiring prior calibration flights or dedicated maneuvers.
\end{abstract}

\input{introduction}
\input{theory_new}
\input{toymodel}
\input{filter_design}
\input{implementation}

\input{conclusion}

\input{acknowledgements}

\clearpage
\appendix
\section*{Appendix}
\input{appendix/tolles_lawson}
\input{appendix/equiv_ekf_ng}
\input{appendix/nn_jacobian}
\input{appendix/crlb_odom}
\input{appendix/jacobian_nn-tl}
\input{appendix/additional_results}

\clearpage
\bibliography{refs}

\end{document}

%% file: src/newcommands.tex
\newcommand{\bfa}{{\bm{a}}}

\newcommand{\bfA}{\bm{A}}
\newcommand{\bfb}{\bm{b}}

\newcommand{\bfB}{\bm{B}}
\newcommand{\hbfB}{\hat{\bm{B}}}

\newcommand{\bfC}{\bm{C}}

\newcommand{\bfF}{\bm{F}}

\newcommand{\bfG}{\bm{G}}

\newcommand{\bfg}{\bm{g}}
\newcommand{\bfh}{\bm{h}}
\newcommand{\bfH}{\bm{H}}

\newcommand{\bfI}{\bm{I}}
\newcommand{\bfJ}{\bm{J}}

\newcommand{\bfK}{\bm{K}}

\newcommand{\bfm}{\bm{m}}

\newcommand{\bfp}{\bm{p}}
\newcommand{\hbfp}{\hat{\bm{p}}}

\newcommand{\bfP}{\bm{P}}

\newcommand{\bfQ}{\bm{Q}}

\newcommand{\bfR}{\bm{R}}

\newcommand{\bfS}{\bm{S}}

\newcommand{\bfu}{\bm{u}}

\newcommand{\bfv}{\bm{v}}

\newcommand{\bfW}{\bm{W}}
\newcommand{\bfw}{\bm{w}}

\newcommand{\bfx}{{\bm{x}}}
\newcommand{\dbfx}{\dot{\bfx}}
\newcommand{\hbfx}{\hat{\bm{x}}}

\newcommand{\bfz}{\bm{z}}

\newcommand{\bobeta}{\bm{\beta}}
\newcommand{\hbobeta}{\hat{\bobeta}}

\newcommand{\boepsilon}{\bm{\epsilon}}

\newcommand{\boomega}{\bm{\omega}}

\newcommand{\bophi}{\bm{\phi}}

\newcommand{\boPhi}{\bm{\Phi}}

\newcommand{\bosigma}{\bm{\sigma}}

\newcommand{\boupsilon}{\bm{\upsilon}}

\newcommand{\boLambda}{\bm{\Lambda}}

\newcommand{\bfzero}{\bm{0}}

\newcommand{\covQ}{\bm{\mathcal{Q}}}
\newcommand{\covR}{\bm{\mathcal{R}}}
\newcommand{\covP}{\bm{\mathcal{P}}}

\newcommand{\diag}{\mathrm{diag}}



%% file: src/preamble.tex
\usepackage{graphicx} 
\usepackage[english]{babel}
\usepackage[utf8]{inputenc}

\usepackage{amsmath}
\usepackage{amssymb}
\usepackage[per-mode = symbol]{siunitx}
\usepackage{xfrac}
\usepackage{bm}

\usepackage{tabularx}
\newcolumntype{Y}{>{\hsize=4\hsize}X}
\usepackage{booktabs}

\usepackage[capitalize]{cleveref}
\crefname{appendix}{App.}{Apps.}
\crefrangeformat{figure}{Figs.~#3#1#4--#5#2#6}
\crefrangeformat{table}{Tabs.~#3#1#4--#5#2#6}
\crefformat{equation}{(#2#1#3)}
\crefrangeformat{equation}{(#3#1#4)-(#5#2#6)}

\usepackage{todonotes}


%% file: src/acronyms.tex
\begin{acronym}
\acro{NN}[NN]{neural networks}
\end{acronym}

\begin{acronym}
\acro{TL}[TL]{Tolles-Lawson}
\end{acronym}

\begin{acronym}
\acro{GNSS}[GNSS]{global navigation satellite system}
\end{acronym}

\begin{acronym}
\acro{MagNav}[MagNav]{Magnetic Anomaly Navigation}
\end{acronym}

\begin{acronym}
\acro{EKF}[EKF]{extended Kalman filter}
\end{acronym}

\begin{acronym}
\acro{KF}[KF]{Kalman filter}
\end{acronym}

\begin{acronym}
\acro{ESKF}[ESKF]{error-state Kalman filter}
\end{acronym}

\begin{acronym}
\acro{RMSE}[RMSE]{root-mean-square error}
\end{acronym}

\begin{acronym}
\acro{CRLB}[CRLB]{Cramér-Rao lower bound}
\end{acronym}


\begin{acronym}
\acro{ML}[ML]{machine learning}
\end{acronym}

\begin{acronym}
\acro{IAE}[IAE]{innovation-based adaptive estimation}
\end{acronym}

\begin{acronym}
\acro{IMU}[IMU]{inertial measurement unit}
\end{acronym}

\begin{acronym}
\acro{RNP}[RNP]{required navigation performance}
\end{acronym}

\begin{acronym}
\acro{ECEF}[ECEF]{Earth-centered Earth-fixed}
\end{acronym}

\begin{acronym}
\acro{FOGM}[FOGM]{First-order Gauss-Markov}
\end{acronym}

\begin{acronym}
\acro{IGRF}[IGRF]{International Geomagnetic Reference Field}
\end{acronym}

\begin{acronym}
\acro{WGS84}[WGS84]{World Geodetic System 1984}
\end{acronym}

\begin{acronym}
\acro{INS}[INS]{inertial navigation system}
\end{acronym}

\begin{acronym}
\acro{RMSE}[RMSE]{root mean square error}
\end{acronym}

\begin{acronym}
\acro{MC}[MC]{Monte-Carlo}
\end{acronym}

\begin{acronym}
\acro{MSE}[MSE]{mean squared error}
\end{acronym}

\begin{acronym}
\acro{FIM}[FIM]{Fisher information matrix}
\end{acronym}

\begin{acronym}
    \acrodef{NG}[NG]{natural gradient}
\end{acronym}

\begin{acronym}
    \acro{DRMS}[DRMS]{distance root mean square}
\end{acronym}

\begin{acronym}
\acro{UTC}[UTC]{Coordinated Universal Time}
\end{acronym}

\begin{acronym}
\acro{UTM}[UTM]{Universal Transverse Mercator}
\end{acronym}

\begin{acronym}
\acro{GPS}[GPS]{Global Positioning System}
\end{acronym}





\begin{acronym}
\acro{FG}[FG]{fluxgate magnetometer}
\end{acronym}


\begin{acronym}
\acro{MSL}[MSL]{mean sea level}
\end{acronym}

%% file: introduction.tex
\section{Introduction}

Airborne \ac{MagNav} aims at using the Earth's crustal magnetic field for positioning and navigation \cite{canciani_absolute_2016, Canciani2017}. Magnetic anomaly field maps can provide a unique geographic fingerprint that serves as a jamming-resistant position reference, complementary to external position aids like GNSS or terrestrial radio signal networks.
The total magnetic field measurements from an aircraft platform-mounted magnetometer are matched to a geo-located map to correct the drift of an inertial navigation system \cite{Hager2025, Canciani2016a}.

A fundamental challenge in \ac{MagNav} is that the total field measured by an aircraft magnetometer is a superposition of multiple sources. The desired signal, originating from crustal anomalies, must be extracted from much larger interference fields generated by the Earth's core \cite{Lesur2022}, the ionosphere \cite{Khazanov2016}, and the aircraft itself \cite{Canciani2022}. With an amplitude of only tens to hundreds of nanoteslas, the navigation signal is typically one to two orders of magnitude weaker than the sources of interference \cite{Canciani2022,gnadt_advanced_2022}. Consequently, accurately compensating for these dominant fields is crucial. Contributions from the Earth's core are reliably removed using regularly updated global models \cite{Alken2021}, whereas ionospheric disturbances can be treated as stochastic noise \cite{canciani_absolute_2016}. The primary difficulty lies in compensating for the aircraft's own magnetic field: interferences from magnetized components as well as from dynamic sources like control surfaces and electric systems \cite{Du2019}.
In dedicated airborne magnetic survey platforms, aircraft-generated interference is minimized by physically separating the magnetometers from the airframe by placing them at the tips of the wings or on a tail stinger. However, such modifications are structurally and regulatorily challenging for commercial aircraft. Hence, the magnetometers must be placed in environments with significant magnetic disturbances, such as within the airframe, which in turn necessitates the use of advanced compensation models and techniques. 
Fields originating from the permanent and induced magnetization of the airframe's nonmoving components (e.g., the fuselage, wings, and structural elements) are typically compensated by the \ac{TL} model and its derivatives, see Appendix~\ref{sec:tolles_lawson} and \cite{Gnadt2022a, liu_application_2022, feng_improved_2022}. 

The \ac{TL} model and its extensions are conventionally applied post-flight to collected sensor data. The static model parameters are calibrated, typically via a least-squares optimization, over an entire flight's dataset.
Offline \ac{TL}-calibration models are still being refined \cite{han_modified_2017}, e.g. by using advanced regression methods \cite{lathrop_accurate_2022, lathrop_magnetic_2023, lathrop_flight_2024} and separately modeling onboard instruments \cite{liu_modified_2024, jukic_applications_2024}.
However, the \ac{TL} model with its rigid-body assumption is fundamentally limited to maneuver-correlated fields. It cannot capture nonlinearities or the dynamic interferences generated by on-board electric currents and moving parts. This limitation motivated the investigation of \ac{ML} approaches and advanced \ac{TL}-\ac{ML} hybrid models \cite{Zhai2023, Laoue2023, Nerrise2024, wang_aeromagnetic_2025, ma_aeromagnetic_2025}.

Although these \ac{ML}-based methods demonstrate improved performance by modeling effects not captured by the \ac{TL} framework, they face significant implementation barriers. First, the requirement for large training datasets creates a logistical bottleneck, as this would require collecting a dedicated dataset for every individual aircraft model to account for unique structural magnetic signatures. Second, the computational complexity of sophisticated methods often exceeds the processing capabilities of standard onboard embedded hardware. Consequently, most contemporary approaches remain offline solutions \cite{Laoue2023, Nerrise2024, wang_aeromagnetic_2025, ma_aeromagnetic_2025, williams_aeromagnetic_1993, Ma2018, yu_aeromagnetic_2022}. This is a critical shortcoming, as it prevents the system from adapting to the dynamic nature of the aircraft's magnetic field that changes continuously due to thermal effects, vibrations, and the operation of onboard systems.

Consequently, there is a growing recognition that adaptive calibration online is crucial for robust performance \cite{dou_novel_2014, dou_adaptive_2021, Canciani2022, beravs_magnetometer_2014, Siebler2020, jiao_real-time_2022}. These systems aim to compensate for changes in interference as they occur during flight.

However, a significant portion of this adaptive calibration literature remains focused on the magnetometer \ac{RMSE} as the sole performance metric, treating compensation as an isolated task. The most relevant measure of success for \ac{MagNav} is, however, the accuracy and integrity of the full navigation solution.
This distinction is crucial because minimizing calibration error is merely an intermediate step and does not inherently guarantee improved positioning accuracy, which also depends on anomaly field characteristics, navigation architecture and filter tuning \cite{Gupta2024, Blakely2025c, Hager2026}. 

The operational viability of such capabilities was recently demonstrated in field trials \cite{Muradoglu2025} using a cold-start system, while the specific algorithmic architecture that enables this has not been published.
%

Ultimately, the ideal solution would be a hybrid online method that is certifiable for operations and adapts in real time without requiring the creation of extensive training datasets. It would leverage a more expressive, non-linear model for maximum accuracy but could revert to a robust linear baseline, like the \ac{TL} model, to ensure operational robustness and prevent unconstrained divergence.
A promising approach in this direction was explored by Gnadt \cite{gnadt_advanced_2022}, combining the \ac{TL} model with a pre-trained \ac{NN}, integrating them into a Kalman-Filter for \ac{MagNav}. The \ac{NN} tracks residual interference not modelled by \ac{TL} in its role as a universal function approximator.
In this \ac{EKF}-based \ac{NN} training \cite{haykin_kalman_2001}, the \ac{NN}'s parameters become part of the filter's state vector, allowing the model to continuously adapt to changing magnetic conditions in real-time.
However, this method still requires both an initialization of the \ac{TL} parameters via a calibration flight and a pre-flight training stage for the \ac{NN} to ``warm start'' the model. 

\input{diagram}

We propose a fully adaptive hybrid online calibration and magnetic anomaly navigation architecture (see \cref{fig:architecture}), featuring
\begin{enumerate}
    \item Flexible Initialization for Operational Readiness (Cold \& Warm Start): We introduce a self-calibrating system that enables ``cold starts'' with zero prior knowledge, removing the bottleneck of dedicated calibration flights. This allows the aircraft to autonomously identify its magnetic signature in-situ, while still supporting ``warm starts'' that leverage parameters from previous flight segments.
    \item Modular, Decoupled Architecture: The architecture integrates the physics-based \ac{TL} model with a \ac{NN} as independent, additive components within the navigation filter, ensuring that the navigation solution remains interpretable and anchored to explainable physical principles.
    \item Data and Computational Efficiency: 
    Integrating both the \ac{TL}- and the \ac{NN}-model parameters as states within the \ac{EKF} allows for fast learning and real-time adaptation of the calibration model parameters alongside the aircraft's position in-flight. This implicit implementation of online Natural Gradient descent for the \ac{NN} parameters provides the geometry-aware rapid convergence properties of a second-order optimizer to model nonlinear interferences. No extensive collection and processing of calibration or training data is required.
\end{enumerate}


We benchmark our methodology against state-of-the-art models using the MagNav Challenge dataset \cite{gnadt_daf-mit_2023}. Our approach advances the hybrid strategies of Gnadt \cite{gnadt_advanced_2022} by demonstrating that our self-calibrating framework achieves comparable navigation accuracy without the prerequisite offline pre-training phase and minimizing reliance on historical flight data.

Our paper is structured as follows. \cref{sec:theory} establishes the theoretical framework, detailing the principles of \ac{KF}-based \ac{NN} training and its connection to standard backpropagation as used in \ac{ML} offline calibration methods. We then introduce this method with a simplified two-dimensional example in \cref{sec:toymodel} to demonstrate how it can be used to model unknown nonlinear interference. Building on this, \cref{sec:filter_design} presents the full filter design for six-degree-of-freedom airborne MagNav, where the magnetic interference of the airborne platform is modeled by our proposed hybrid \ac{TL} and \ac{NN} architecture. In \cref{sec:implementation}, the algorithm's  magnetic field compensation and positioning performance is validated on the \ac{MagNav} challenge dataset \cite{gnadt_daf-mit_2023} dataset. 

%% file: diagram.tex
\definecolor{physicsblue}{RGB}{220, 230, 255}
\definecolor{nngreen}{RGB}{220, 255, 220}
\definecolor{filtergray}{RGB}{245, 245, 245}
\definecolor{sensorred}{RGB}{255, 235, 235}
\definecolor{processyellow}{RGB}{255, 250, 230}

\begin{figure}
    \centering
    \resizebox{\linewidth}{!}{%
    \begin{tikzpicture}[
        >={Latex[width=2mm,length=2mm]},
        node distance=1.5cm and 2.0cm,
        auto,
        font=\footnotesize,
        sensor/.style={draw, fill=sensorred, rectangle, minimum height=2.5em, minimum width=6em, drop shadow, rounded corners, align=center, font=\footnotesize\bfseries},
        process/.style={draw, fill=processyellow, rectangle, minimum height=3em, minimum width=7em, drop shadow, rounded corners, align=center},
        statebox/.style={draw, fill=filtergray, rectangle, minimum height=4em, minimum width=11em, drop shadow, rounded corners, align=center},
        model/.style={draw, fill=white, rectangle, minimum height=3em, minimum width=10em, drop shadow, rounded corners, align=center},
        sum/.style={draw, fill=white, circle, inner sep=2pt},
        arrow/.style={draw, ->, thick},
        line/.style={draw, thick},
    ]
    
        
        \node [sensor] (imu) {IMU \\ Acc $\bfa$, Gyro $\boomega$};
        \node [process, right=1.5cm of imu] (ins) {Strapdown INS \\ $\int$ };
        
        \node [sum, right=1.5cm of ins] (pos_sum) {$\Sigma$};
        \node [above=1.0cm of pos_sum, font=\bfseries] (output_label) {Corrected Position $\hat{p}$};
        \draw [dashed, ->] (pos_sum) -- (output_label);
        
        \node [model, right=1cm of pos_sum] (map) {
            \textbf{Global Models} \\ 
            $B_{\text{core}}(\hat{\bfp}) + B_{\text{anom}}(\hat{\bfp})$
        };
    
        
        \node [statebox, below=0.75cm of ins, text width=3.5cm] (state) {
            \textbf{Augmented EKF State} $\hbfx$ \\[0.3em]
            \scriptsize
            Nav Errors $\delta \hbfx$ \\
            TL Params $\hat{\beta}_{\text{TL}}$ \\
            NN Params $\hat{\Lambda}_{\text{NN}}$
        };
        
    
        \node [model, fill=physicsblue, below=2.8cm of map, yshift=1.2cm] (tl) {
            \textbf{Tolles-Lawson} \\ 
            $f_{\text{TL}}(\bfm, m, \hat{\beta}_{\text{TL}})$
        };
    
        \node [model, fill=nngreen, below=1.0cm of tl] (nn) {
            \textbf{Neural Network} \\ 
            $\tilde{g}_{\text{res}}(\bfm, m, \hat{\Lambda}_{\text{NN}})$
        };
    
        \coordinate (mid_tlnn) at ($(tl)!0.5!(nn)$);
        
        \coordinate (mag_center) at (imu |- mid_tlnn);
        
        \def\magsep{1.0cm} 
        
        \node [sensor] (vecmag) at ($(mag_center)+(0,\magsep)$) {Vector Mag \\ $\bfm$};
        \node [sensor] (scalmag) at ($(mag_center)+(0,-\magsep)$) {Scalar Mag \\ $m$};
    
        \node [right=0.5cm of nn, font=\scriptsize, draw, inner sep=2pt] (scale) {$\times \alpha$};
    
    
        \node [sum, right=1.2cm of map] (sum_total) {$\Sigma$};
        
        \node [sum] (sum_int) at (sum_total |- tl) {$\Sigma$}; 
        
        \node [sum, right=1.2cm of sum_total] (innov) {$\Sigma$};
    
        \node [process, fill=filtergray, below=5.5cm of innov, text width=2.5cm] (ekf) {
            \textbf{EKF Update} \\ 
            $K \cdot \nu_t$ \\
            Correct State
        };
    
    
        \draw [arrow] (imu) -- (ins);
        \draw [arrow] (ins) -- node[above, font=\scriptsize] {$\mathbf{p}_{\text{ins}}$} (pos_sum);
        \draw [arrow] (pos_sum) -- (map);
        \draw [arrow] (map) -- node[midway, above, font=\scriptsize] {$B_{\text{ext}}$} (sum_total);
    
        \draw [arrow, dashed] (ins.south) -- node[pos=0.35, right, font=\scriptsize] {$\hat{\mathbf{x}}_{\text{ins}}$} (state.north);
    
        \draw [arrow] (state.east) 
            -| node[pos=0.75, right, font=\scriptsize] {$\delta \hat{\mathbf{p}}$} (pos_sum.south);
    
        
            
    
        \draw [line, dashed] (vecmag.south) -- (mag_center);
        \draw [line, dashed] (scalmag.north) -- (mag_center);
        
        \draw [line, dashed] (mag_center) --
            node[near end, above, font=\scriptsize] {$\bfm,\, m$}
            (mid_tlnn);
        
        \draw [arrow, dashed] (mid_tlnn) -| (tl.south);
        \draw [arrow, dashed] (mid_tlnn) -| (nn.north);
    
            
        
    
        \coordinate (param_bus) at (pos_sum |- state);    
        
        \draw [arrow] (state.east)
            |- (param_bus)
            |- node[near end, above, font=\scriptsize] {$\hat{\beta}_{\text{TL}}$}
               (tl.west);
        
        \draw [arrow] (state.east)
            |- (param_bus)
            |- node[near end, above, font=\scriptsize] {$\hat{\Lambda}_{\text{NN}}$}
               (nn.west);
    
    
        \draw [arrow] (tl.east) -- (sum_int);
        \draw [arrow] (nn.east) -- (scale.west);
        \draw [arrow] (scale.east) -| (sum_int.south);
        
        \draw [arrow] (sum_int.north) -- node[midway, right, font=\scriptsize] {$B_{\text{pf}}$} (sum_total.south);
        
        \draw [arrow] (sum_total.east) -- node[above, font=\scriptsize] {$h(\hat{x})$} (innov.west);
    
        \draw [arrow] (innov.south) -- (ekf.north);
        \draw [arrow] (ekf.west) -| ($(state.south)+(0,-1.0)$) -| (state.south);
    
        \node [draw=black!40, dashed, fit=(tl) (nn) (sum_int) (scale), inner sep=0.3cm, label=above:\textit{NN-augmented calibration model}] {};
    
    \end{tikzpicture}}
    \caption{Hybrid MagNav Architecture with \ac{NN}-augmented calibration model.}
    \label{fig:architecture}
\end{figure}
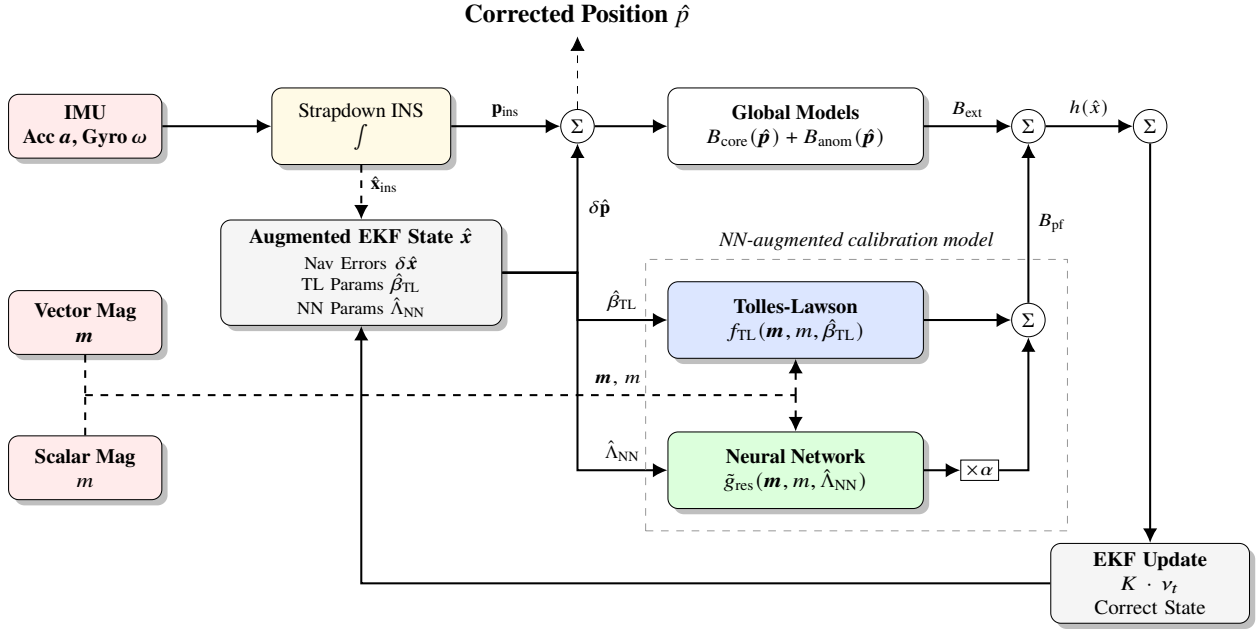

%% file: theory_new.tex
\section{A Unified Framework for State Estimation and Model Learning}
\label{sec:theory}

Hybrid \ac{EKF}-\ac{NN} systems are promising for navigation, as the methods' strengths complement each other. The performance of an \ac{EKF} is sensitive to model inaccuracies and unknown noise (i.e., linearization errors, map inaccuracies, or platform calibration errors), while \ac{NN}s can learn complex, non-linear interference dynamics from data without prior models. This combination leverages the \ac{EKF} efficiency with \ac{NN} data-driven robustness for model identification.

NNs are typically integrated with \ac{EKF}s by either performing specific tasks, like predicting \ac{INS} errors during \ac{GNSS} outages to correct state estimates \cite{wang_neural_2007, guo_neural_2013}, or by adaptively tuning the filter's parameters, such as its noise covariance matrices or the \ac{KF} gain \cite{jouaber_nnakf_2021, revach_kalmannet_2022, Cuenca2024, cohen_inertial_2024}.

Our proposed integration strategy augments the \ac{EKF} state vector with the \ac{NN}'s parameters (weights and biases), enabling their continuous, real-time adaptation. The \ac{EKF} updates the \ac{NN} to model unknown nonlinear interference signals on the measurement simultaneously to system state estimation without extensive prior knowledge. The filter leverages the information geometry of the parameter space to achieve learning speed and convergence robustness that outperforms standard first-order backpropagation.
Our architecture offers distinct advantages for airborne applications:
\begin{enumerate}
    \item Suited for Online Learning: Unlike classical recurrent \ac{NN} architectures (e.g., LSTMs), our framework requires no pre-training or data buffering and is inherently suited for real-time operations
    \item Efficient Convergence: By incorporating second-order information via the \ac{NG}, the optimization converges significantly faster than first-order methods, beneficial for adapting to dynamic conditions during flight
    \item Embedded Hardware Compatibility: The compact \ac{NN} architecture maintains low state dimensionality, enabling full-covariance updates of the parameters on standard embedded hardware without the prohibitive computational costs typically associated with second-order optimization.
    \item Certifiability: Relying on the well-established probabilistic framework of the \ac{KF} is a key advantage for high-integrity applications. This grounds the continuous learning process in a deterministic framework widely accepted in aviation, in contrast to heuristic-based optimization methods.
\end{enumerate}
 
\subsection*{Application to Magnetic Anomaly Navigation with Simultaneous Platform Calibration} 
We apply this hybrid EKF-NN framework to the simultaneous navigation and online calibration for airborne \ac{MagNav}.
We augment the standard \ac{EKF} state vector, comprising navigation and magnetometer states, with the \ac{NN} parameters, $\bfx_{\text{NN}}$, that learn to compensate for the platform's magnetic interference.  In this context, calibration is the operational objective of removing the aircraft's magnetic field from the measurements, the \ac{NN} being part of the calibration model, while learning is the algorithmic mechanism used to achieve it.

This joint \ac{KF} approach enables the learning of the \ac{NN} in a recurrent fashion, where the filter's covariance matrix provides the memory. Our framework implicitly computes the state's gradient with respect to the parameters and uses this information to perform an online Natural Gradient update on the parameters \cite{Ollivier2018}. This enables true RTRL (Real-Time Recurrent Learning) model learning during operations, unlike classical Recurrent Neural Networks (RNNs) that typically rely on computationally expensive BPTT (Backpropagation Through Time) on buffered sequences.

The \ac{NN} becomes part of the filter's observation model, predicting the highly nonlinear components of the magnetic interference generated by aircraft systems. During the \ac{EKF}'s correction step, the magnetic innovation error (the difference between measured and predicted fields) is used to update the entire augmented state simultaneously.
The results is a symbiotic process where the continuous learning of the interference model improves the navigation solution, and a better navigation solution provides a cleaner signal for refining the calibration parameters.
The filter's recursive update mechanism itself trains the network, a process analogous to a natural gradient descent. Temporal information is inherently encoded, removing the need to manually add features like measurement derivatives.

This simultaneous navigation and system identification approach is first demonstrated in a conceptual magnetometer-aided odometry model in the following chapter, and then experimentally applied to airborne \ac{MagNav}.
\subsection*{Extended Kalman Filter for Training Neural Networks}

The training of a \ac{NN} involves optimizing its weights and biases, a task predominantly accomplished through gradient descent-based algorithms. The most ubiquitous of these is the backpropagation algorithm \cite{Rumelhart1986, Schmidhuber2015}. An alternative paradigm employs classical recursive estimation techniques, such as the \ac{KF} \cite[Ch.~4.2]{BrownHwang2012}, which frames the learning process as an optimal state estimation problem. Although conceptually distinct, these two methods are fundamentally linked.

The connection is made through the lens of information geometry and the \ac{NG}. Standard backpropagation (i.e., ordinary gradient descent) follows the direction of steepest descent in the Euclidean parameter space. In contrast, the \ac{NG} follows the direction of steepest descent on the statistical manifold defined by the model's probability distributions, where distances are measured using the Kullback-Leibler (KL) divergence. It achieves this by preconditioning the gradient with the inverse of the Fisher Information Matrix (FIM) that locally characterizes the curvature of the distribution space. As a result, \ac{NG} updates are invariant to the parameterization of the model and, under some assumptions, provide the most efficient direction for learning in terms of information geometry \cite{Amari1998}.

The connection to Kalman filtering is fundamental: Applying an \ac{EKF} to estimate a fixed, unknown parameter of a probabilistic model (such as a neural network's parameters) is under the Gaussian noise assumption equivalent to performing an online, stochastic \ac{NG} descent on the log-likelihood of the observations \cite{Ollivier2018}. This equivalence was later generalized to the full dynamical system, proving that the standard \ac{EKF} performs \ac{NG} descent in the abstract space of system trajectories \cite{Ollivier2019}.

To formalize this methodology, we first recap the standard gradient descent approach. As notation conventions often differ, we adopt one that aligns closely with Kalman filtering to ensure consistency.

Let $\bfh(\bfx_{\text{NN}})$ represent the network's output vector for a given parameter/state vector $\bfx_{\text{NN}}$ (weights and biases) and training data set, and let $\bfz$ be the corresponding target vector. The goal is to minimize a loss function, typically the Mean Squared Error (MSE), that is, for a single sample: 
\begin{align} 
    L = \frac{1}{2} (\bfz - \bfh(\bfx_{\text{NN}}))^\top(\bfz - \bfh(\bfx_{\text{NN}})) 
    \label{eq:loss_function}
\end{align}
The standard gradient descent algorithm with backpropagation updates the parameters using backpropagation, i.e. moving along the negative gradient of this loss. Neglecting more advanced terms like momentum, the basic update rule is:
\begin{align}
\Delta \bfx_{\text{NN}} (t) &= -\eta \frac{\partial L(t)}{\partial\bfx_{\text{NN}}} = \eta \left( \frac{\partial \bfh(\bfx_{\text{NN}})}{\partial \bfx_{\text{NN}}} \right)^\top (\bfz - \bfh(\bfx_{\text{NN}})) 
\label{eq:delta_weights_nn}
\end{align}
We now frame the network training process using an \ac{EKF}. The network's parameters $\bfx_{\text{NN}}$ are treated as the filter's state vector. The weight matrices $\bfW$ and bias vectors $\bfb$ of a network with $L$ layers are flattened into the column vector
 \begin{align}
    \bfx_{\text{NN}} = \left[ 
    \text{vec}(\bfW^{(1)}), 
    \bfb^{(1)}, 
    \dots, 
    \text{vec}(\bfW^{(L)}), 
    \bfb^{(L)} 
 \right]^\top.
 \end{align}
Ideally, the optimal parameters are time-invariant, suggesting a static process $\dot{\bfx}_{\text{NN}} = 0$.
However, a purely static model is not only unrealistic but also problematic, as the state covariance matrix $\covP$ shrinks with new data, causing the \ac{KF} gain $\bfK$ to vanish and eventually stop the learning. A static \ac{EKF} with no process noise ($\covQ_c=0$) is equivalent to an online \ac{NG} descent with a decaying learning rate $\eta_t = 1/(t+1)$ \cite[~Th.2]{Ollivier2018}.
To keep the filter adaptive, process noise 
is introduced. This creates a ``fading memory'' filter, equivalent to an \ac{NG} descent with a larger or constant (non-decaying) learning rate. The process model is
\begin{align}
\dot{\bfx}_{\text{NN}}(t) &= \bfw(t), \quad \text{where } \bfw(t) \sim \mathcal{N}(0, \sigma_{\bfw}^2 \bfI)
\end{align}
For implementation in discrete time, the process noise covariance, $\covQ_d$, can be discretized from the spectral density $\covQ_c := \sigma_{\bfw}^2 \bfI$ via a simple first-order approximation ($\covQ_d \approx \covQ_c \Delta t$) or more advanced methods that provide higher accuracy.
The discrete-time measurement model uses the network's output as the predicted measurement. The relationship between the true output $\bfz_t$ and the network's prediction $\bfh(\bfx_{\text{NN}, t})$ is given by:
\begin{align}
\bfz_t = \bfh(\bfx_{\text{NN},t}) + \boupsilon_t, \quad \text{where } \boupsilon_t \sim \mathcal{N}(0, \sigma_{\boupsilon}^2 \bfI)
\end{align}
with measurement noise covariance $\covR:= \sigma_{\boupsilon}^2 \bfI$. 

The \ac{EKF} training process then consists of the standard prediction-correction loop. For a static model with added noise, the steps are:

\begin{enumerate}
    \item Prediction (Time Update):
    \begin{align}
    \bfx_{\text{NN},t|t-1} &\leftarrow \bfx_{\text{NN},t-1} \\
    \covP_{t|t-1} &\leftarrow \covP_{t-1} + \covQ_d
    \end{align}
    \item Correction (Measurement Update):
    \begin{subequations}
    \begin{align}
    \bfK_t &= \covP_{t|t-1} \bfH^\top (\bfH \covP_{t|t-1} \bfH^\top + \covR)^{-1} \\
    \bfx_{\text{NN},t} &= \bfx_{\text{NN},t|t-1} + \bfK_t (\bfz_t - \bfh(\bfx_{\text{NN},t|t-1})) \label{eq:kf_update_state} \\
    \covP_t &= (\bfI - \bfK_t \bfH) \covP_{t|t-1} \label{eq:ekf_update_covariance},
    \end{align}
    \label{eq:ekf_update}
    \end{subequations}
\end{enumerate}
where $\bfH$ is the Jacobian of the network output $\bfh$ with respect to the parameters $\bfx_{\text{NN}}$, evaluated at the current state estimate $\hbfx_{\text{NN}}$.

The update to the network parameters in \cref{eq:kf_update_state}, $\Delta \bfx_{\text{NN},t|t-1} = \bfK_t (\bfz_t - \bfh(\bfx_{\text{NN}, t|t-1}))$, is, as demonstrated in \cite{Ollivier2018} and briefly recapped in Appendix~\ref{ssec:equivalence_ekf_ng}, conceptually equivalent to an online, stochastic \ac{NG} descent. 
Under the assumption of a Gaussian observation model, the inverse of the \ac{EKF}'s covariance matrix, $\covP^{-1}$, approximates the \ac{FIM} $\bfJ$. The \ac{EKF} update of the state is an online gradient descent pre-conditioned by $\bfJ^{-1}$, the very definition of a \ac{NG}.
With this, the standard backpropagation rule \cref{eq:delta_weights_nn} is revealed as the most simplified case of the \ac{EKF}:
\begin{enumerate}
    \item \ac{EKF} (Full $\covP$) $\approx$ Full-Matrix \ac{NG}: Tracks the inverse Fisher information matrix and performs a second-order update following the steepest descent on the probabilistic manifold, scales quadratically, $\mathcal{O}(N^2)$.
    \item Diagonal \ac{EKF} ($\covP = \diag(\covP_{ii})$) $\approx$ Diagonal \ac{NG}: Assumes parameters are decoupled and uses a per-parameter scaling. This is analogous to the second-moment adaptation in modern optimizers like Adam and RMSprop and reduces complexity to $\mathcal{O}(N)$ .
    \item Isotropic \ac{EKF} ($\covP = p\bfI$) $\approx$ Ordinary Gradient Descent (Backpropagation): Assuming $\covP$  and $\covR = \epsilon^{-1}\bfI$ are isotropic causes the \ac{KF} gain $\bfK_t$ to reduce to a scalar multiple of $\bfH^\top$, scaling with $\mathcal{O}(N)$. In the special case of a constant scalar gain, we recover the standard backpropagation parameter update \cref{eq:delta_weights_nn} \cite{ruck_comparative_1992, sarat_chandran_comments_1994}.
\end{enumerate}

In deep learning, the full-matrix \ac{EKF} (Option 1) is typically computationally prohibitive, necessitating diagonal approximations (Option 2). However, our application targets a shallow network architecture with few parameters. This places us in a computationally favorable regime where the full covariance matrix $\covP$ can be tracked explicitly. The full-matrix \ac{EKF} captures all parameter correlations, making the optimization significantly more robust to initialization than standard diagonal methods and converging to a steady-state regime in fewer iterations.
\subsection*{Tuning the Learning Rate}
\label{ssec:tuning_theory}
The equivalence of \ac{EKF} and \ac{NG} provides a framework for automatically adjusting the learning rate of the network by controlling the \ac{EKF}'s covariance matrices. While the process noise covariance $\covQ_d$ and measurement noise covariance $\covR_d$ are tuned as static hyperparameters, the resulting \ac{KF} gain $\bfK_t$ remains dynamic. As detailed in Appendix~\ref{sssec:ekf_ng_learningrates}, this has two phases. The initial learning rate is dictated by the initial covariance, $\covP_0$ that acts as our Bayesian prior on the parameters. The effective learning rate then gets naturally annealed down to a steady-state regime determined by the $\covQ/\covR$ ratio, allowing the network to converge fast and then maintain continuous adaptation.

%% file: toymodel.tex
\section{Conceptual Model: Magnetic-Anomaly Aided Odometry}
\label{sec:toymodel}
In this chapter, we present a conceptual model to demonstrate the core principle of simultaneous online system calibration and state estimation for navigation.
In aided navigation problems, the sensor measurements are often not only corrupted by simple white noise, but also encompass colored noise and interference components with a nonlinear relation to the state.
When a platform-mounted magnetometer measurement is corrupted by electric currents and moving magnetic parts, a fixed measurement noise covariance matrix $\covR$ is insufficient to model the platform interference fields that need to be removed from the measurement. The interference model parameters must be determined and adapted in real-time to maintain accuracy.

To illustrate this, we set up an \ac{EKF} for magnetic-anomaly aided odometry in a 2-D plane, where the platform's magnetic interference is a highly nonlinear, state-dependent function. The odometry state vector is augmented to include the parameters of a platform magnetic interference model.
We consider two cases of increasing complexity:\\
Scenario 1: Assuming that the mathematical structure of the nonlinear interference model is known, the filter estimates the model parameters online. This establishes a baseline for the state augmentation approach without the complexity of a neural network and is an analogue to Online Calibration for \ac{MagNav} using the \ac{TL}-model \cite{Canciani2022}.\\
Scenario 2: The filter has no prior knowledge of the interference function and must train a simple neural network to learn it from scratch during navigation.

\subsection{EKF for joint state and interference model parameter estimation}
\label{ssec:system_model}

We consider a vehicle moving in a 2-D plane with a variable velocity around \SI{20}{\meter\per\second} whose state is estimated sequentially. The system's primary state is its cartesian position $\bfx_t = \left[p_x ,\,\, p_y \right]^\top$. Its dynamics are governed by the discrete-time linear model
\begin{equation}
    \bfx_t = \bfF \bfx_{t-1} + \bfG \bfu_{t-1} + \bfw_{t-1}
    \label{eq:cv_model_dynamics}
\end{equation}
with the state transition matrix $\bfF = \bfI_{2x2}$, control input matrix $\bfG = \Delta t\bfI_{2x2}$, velocity control input $\bfu_{t-1} = \left[v_x ,\,\, v_y \right]^\top \in \mathbb{R}^2$, and Gaussian process noise  $\bfw \sim \mathcal{N}(\bfzero, \bfQ_c)$.
The vehicle receives scalar measurements of the local magnetic field intensity $z_t$ at each timestep. The measurement is a function of the true magnetic anomaly field $m(\bfx_t)$, the nonlinear platform field $g(\boldsymbol{\phi}_t)$, and uncorrelated, zero-mean Gaussian noise:
\begin{equation}
    z_t = m(\bfx_t) + g(\boldsymbol{\phi}_t) + \upsilon_t, \quad \upsilon_t \sim \mathcal{N}(0, \sigma_{\boupsilon}^2)
\end{equation}
We define the ground truth of the platform interference field, $g(\cdot)$, as a nonlinear function whose inputs are drawn from the full feature vector $\boldsymbol{\phi}_t$. The features are chosen to represent physical quantities known to directly influence a platform's magnetic field in \ac{MagNav}. The full feature vector at timestep $t$ is $\bophi_t = [p_{x,t}, p_{y,t}, m(\bfx_t), \psi_t, s_{t}]^\top$, where $\psi_t$ represents the vehicle's heading (here assumed coincident with the track direction) and $s_t = \sqrt{v_{x}^2 + v_{y}^2}$ is its speed. The coefficients $(\beta_1, \dots, \beta_7)$ are constants that define the shape of our arbitrarily chosen ground truth function:
\begin{equation}
\begin{aligned}
    g(\boldsymbol{\phi}_t) = \beta_1 \sin(\beta_2 p_{x,t}) + \beta_3 \cos(\beta_4 p_{y,t}) + \beta_5 \left(\frac{m(\bfx_t)}{c}\right)^2
    + \beta_6 \cos(\psi_t) + \beta_7 v_{\text{tot},t}^2, \quad  c \in \mathbb{R}
\end{aligned}
\label{eq:noise_gt}
\end{equation}
The core challenge is that the function $g(\cdot)$ is unknown to the filter.

In scenario 1, the filter knows the mathematical structure of $g(\cdot)$ and must only estimate its parameters, $\boLambda_{\bobeta} = [\beta_1,\, \dots,\, \beta_7]^\top$.
In the more realistic scenario 2, the filter does not know the function's structure. It must approximate it using a neural network, denoted $\tilde{g}(\hat{\boldsymbol{\phi}}_t, \bfx_{\text{NN}})$. The $N_p$ unknown/adaptive weights and biases are grouped into the parameter vector $\boLambda_{\text{NN}} = \bfx_{\text{NN}} \in \mathbb{R}^{N_p}$ as known from our theoretical derivation in section \cref{sec:theory}.
Because we are interested in the filter's performance with incomplete information on the magnetic interference sources, the network input is restricted to a subset of the true features used to generate the interference ground truth \cref{eq:noise_gt}, $\bophi_{\text{i},t} \subseteq \bophi_t$.
This reduces model complexity, avoids correlated inputs, and simplifies the hardware architecture.

To simultaneously estimate the vehicle's position and the interference model parameters $\boLambda$, we augment the state vector to $\bfx_{\text{aug}, t}  = [\bfx_t^\top,\, \boLambda_t^\top ]^\top \in \mathbb{R}^{2+N_p}$.
The dynamics of the model parameters $\boLambda$ are modeled as a random walk in both scenarios. This assumes the parameters are nearly constant, but allows the filter to adapt them over time. This leads to the following EKF formulation.

\paragraph*{Prediction}
The primary system state is predicted using \cref{eq:cv_model_dynamics}, while the interference model parameters remain unchanged:
\begin{equation}
    \hat{\bfx}_{\text{aug}, t|t-1} = \begin{bmatrix} \bfF\hat{\bfx}_{t-1|t-1} + \bfG \bfu_{t-1} \\ \hat{\boLambda}_{t-1|t-1} \end{bmatrix} + 
    \begin{bmatrix}
        \bfw_{x,t-1} \\ \bfw_{\Lambda,t-1}
    \end{bmatrix}
\end{equation}
The state covariance matrix $\covP$ is propagated using the augmented process noise $\covQ_{d, \text{aug}}$, comprising the discretized vehicle's process noise $\covQ_d = \bosigma^2_{\bfw_x} \bfI$ and a small $\covQ_{\boLambda} = \bosigma^2_{\bfw_{\Lambda}} \bfI := q_{\boLambda}\bfI$ for the parameters $\boLambda$ to allow for learning.
\begin{equation}
    \covP_{t|t-1} = \boPhi \covP_{t-1|t-1} \boPhi^\top + \covQ_{d, \text{aug}}, \quad \text{where} \quad \boPhi = \begin{bmatrix} \bfF & \bfzero \\ \bfzero & \bfI \end{bmatrix}, \, \covQ_{d, \text{aug}} = \begin{bmatrix} \covQ_d & \bfzero \\ \bfzero & q_{\boLambda}\bfI \end{bmatrix}
\end{equation}

\paragraph*{Update}
\label{ssec:toymodel_update}
The standard EKF update \cref{eq:ekf_update} differs only in the predicted measurement, $h(\hat{\bfx}_{\text{aug}, t|t-1})$ for the two scenarios.

\textit{Scenario 1:} Known nonlinear function structure $g(\cdot)$ with the current parameter estimates:
\begin{equation}
    h(\hat{\bfx}_{\text{aug}, t|t-1}) = m(\hat{\bfx}_{t|t-1}) + g(\hat{\bophi}_t, \hat{\boLambda}_{\bobeta,t|t-1})
    \label{eq:measurement_eq_parameterized_model}
\end{equation}

\textit{Scenario 2:} The neural network approximates the nonlinear platform interference $\tilde{g}(\cdot)$:
\begin{equation}
    h(\hat{\bfx}_{\text{aug}, t|t-1}) = m(\hat{\bfx}_{t|t-1}) + \tilde {g}(\hat{\bophi}_{\text{i},t}, \hat{\boLambda}_{\text{NN},t|t-1})
\end{equation}

\paragraph*{Measurement Jacobian}
\label{sssec:jacobian}
For the first scenario, where the analytical form of the platform interference function is known, the derivation of the Jacobian $\bfH_t$ is straightforward. For the second scenario, the Jacobian contains the derivatives of the \ac{NN} with respect to its input and parameters:
\begin{equation}
    \bfH_t = \frac{\partial h}{\partial \bfx_{\text{aug}}} \bigg|_{\hat{\bfx}_{\text{aug}, t|t-1}} = \begin{bmatrix} \left(\frac{\partial m}{\partial \bfx} + \frac{\partial \tilde{g}}{\partial \bfx}\right)\bigg|_{\hat{\bfx}_{\text{aug}, t|t-1}} & \frac{\partial \tilde{g}}{\partial \boLambda_{\text{NN}}}\bigg|_{\hat{\bfx}_{\text{aug}, t|t-1}} \end{bmatrix}.
    \label{eq:cv_model_nn_jacobian}
\end{equation}
The term $\frac{\partial \tilde{g}}{\partial \bfx}$ is generally non-zero because the network's input features $\bophi_{\text{i}}$ can depend on the vehicle's state $\bfx$ (e.g., using position or the map value at that position as a feature), but we will neglect these contributions. The calculation of the \ac{NN} Jacobian is standard and illustrated in Appendix~\ref{sec:nn_jacobian}.

\subsection{Results}
\label{ssec:results}

We present the simulation results for the two scenarios outlined in \ref{ssec:toymodel_update}. 
The simulations were run on a standard consumer-grade CPU using MATLAB and the Deep Learning Toolbox.
The reported performances and runtimes are averaged over 100 independent \ac{MC} trials for each network configuration to ensure statistical reliability.

\subsubsection{Scenario 1: Known, Parameterized Interference Model}
\label{sssec:results_known_form}

The magnetic-anomaly aided odometry for a filter with perfect knowledge of the nonlinear interference model's structure and estimates the model parameters $\bobeta$ converges reliably. The mean position error over 100 \ac{MC} runs is \SI{0.72}{\meter} (standard deviation of 0.01), showing that the filter robustly binds the average odometry drift of \SI{21}{m} (standard deviation \SI{8}{\meter}). A low \ac{RMSE} of the model approximation (\SI{0.63}{\nanotesla}, standard deviation \SI{0.01}{\nanotesla}) confirms that the filter is able to accurately estimate the true parameters of the interference model online. 

\subsubsection{Scenario 2: Unknown Interference modeled by a Neural Network}
\label{sssec:results_nn}
\begin{figure}[tbh!]
    \centering
    \includegraphics[width=\textwidth]{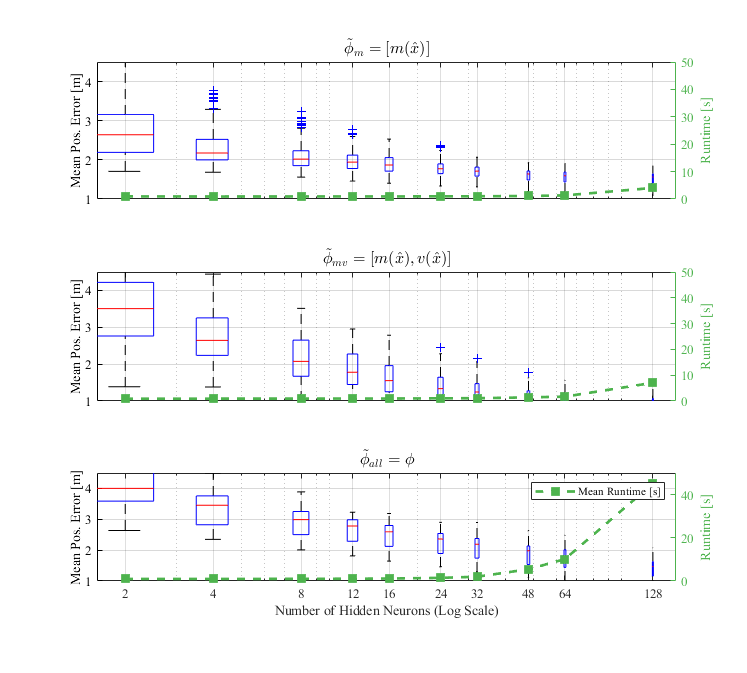}
    \caption{Mean position error distribution and average runtimes from 100 Monte Carlo simulations for input parameter sets $\bophi_{\text{m}}$, $\bophi_{\text{mv}}$, and $\bophi_{\text{all}}$ for different numbers of hidden neurons.}
    \label{fig:mc_summary_sweep}
\end{figure}

The neural network architecture was designed for computational efficiency, consisting of an input layer, a single hidden layer with a $\mathrm{tanh}(\cdot)$ (hyperbolic tangent) activation function, and a single output neuron with a linear activation to produce the scalar interference estimate. This shallow, fully-connected \ac{NN} was tested in configurations using 2 to 128 hidden neurons $N_h$, and three different feature sets (magnetometer only $\bophi_{\text{m}}$, magnetometer and velocity $\bophi_{\text{mv}}$, and all inputs to the true nonlinear function $\bophi_{\text{all}} = \bophi$). The \ac{NN} parameters were initialized using the Glorot initialization \cite{glorot_understanding_2010}, and the \ac{EKF} covariances as $\covP_{0} = \mathrm{diag}(\bfI_{2 \times 2}, 1000 \bfI_{N_p \times N_p})$, $\covQ_d = \mathrm{diag}( 20 \bfI_{2 \times 2} , 0.5 \bfI_{N_p \times N_p} )$, and $\covR = 0.1^2$. \cref{fig:mc_summary_sweep} shows the mean error and its distribution for this hyperparameter sweep, along with the averaged runtime.

The results for this more challenging scenario are shown in \cref{fig:mc_summary_sweep} and highlight a trade-off between computational cost and robust performance. Generally, increasing the number of neurons improves the consistent performance of the adaptive system, leading to lower median error and more narrow error distribution, but at the cost of longer computational runtime. 
The accuracy loss for small neuron numbers can, however, be tolerable when small state spaces have to be prioritized. The reduced feature sets $\bophi_{\text{m}}$ and $\bophi_{\text{mv}}$ consistently yield better results than the full feature set $\bophi_{all}$. The minimal configuration using only the magnetometer as input demonstrates robust performance, achieving position errors only marginally higher than when additional velocity information is used, at lower interquartile ranges but with more outliers. For very low neuron numbers, its median even outperforms the other configurations. 
Interestingly, providing all available inputs degrades performance, likely due to feature redundancy and the increased challenge of learning from a higher-dimensional input space with a small network.

An analysis of the Cramér-Rao Lower Bound (CRLB) confirms that increasing network complexity without additional information dilutes system observability, fundamentally raising the theoretical uncertainty floor for the navigation states (see Appendix~\ref{sec:CRLB_suppl} for details).

\paragraph*{Detailed Analysis of a Representative Case}
To understand how the filter achieves these results, we examine a single, representative simulation run for each trajectory with $N_h = 3$ and the minimal feature set $\bophi_{\text{m}}$.
\begin{figure}[tbh!]
    \centering
    \includegraphics[width=\textwidth]{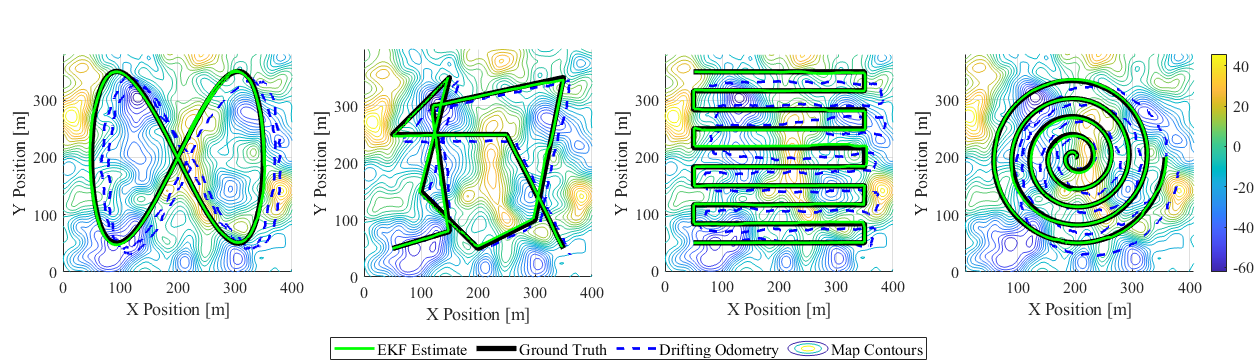}
    \caption{Magnetic-Anomaly aided odometry with \ac{NN}-online-learning based platform interference compensation for four trajectories ($\bophi = \bophi_{\text{m}}$, $N_h = 8$).}
    \label{fig:traj_comparison}
\end{figure}

\begin{figure}[tbh!]
    \centering
    
    \begin{subfigure}[b]{0.48\textwidth}
        \centering
        \includegraphics[width=\textwidth]{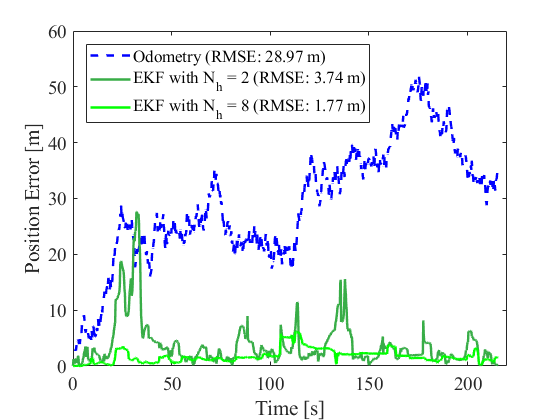}
        \caption{}
        \label{fig:position_error_lawnmower}
    \end{subfigure}
    \hfill 
    \begin{subfigure}[b]{0.48\textwidth}
        \centering
        \includegraphics[width=\textwidth]{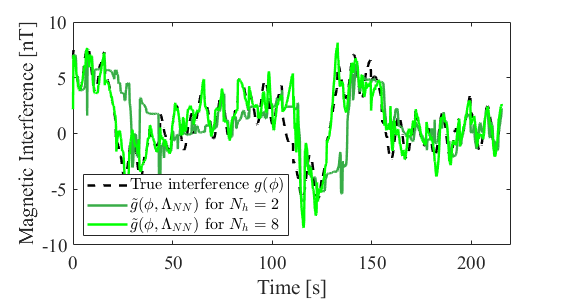}
        \caption{}
        \label{fig:calibration_error_lawnmower}
    \end{subfigure}
    
    \caption{NN-EKF performance on the lawnmower trajectory ($\bophi_{\text{m}}$) for $N_h = 2$ and $N_h = 8$ showing position RMSE improvement compared to odometry in (a) and platform interference approximation in (b).}
    \label{fig:lawnmower_performance} 
\end{figure}

\begin{figure}[tbh!]
    \centering
    
    \begin{subfigure}[b]{0.48\textwidth}
        \includegraphics[width=\textwidth]{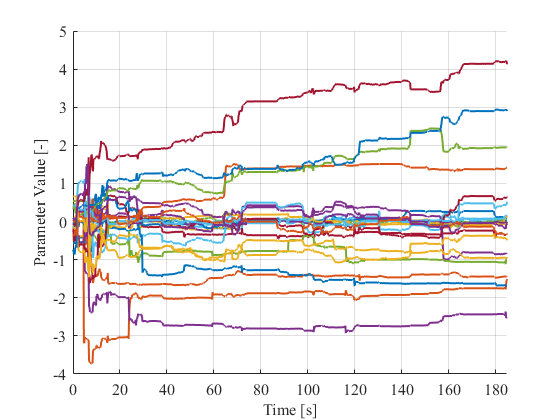}
        \caption{}
        \label{subfig:param_fig_eight}
    \end{subfigure}
    \hfill 
    \begin{subfigure}[b]{0.48\textwidth}
        \includegraphics[width=\textwidth]{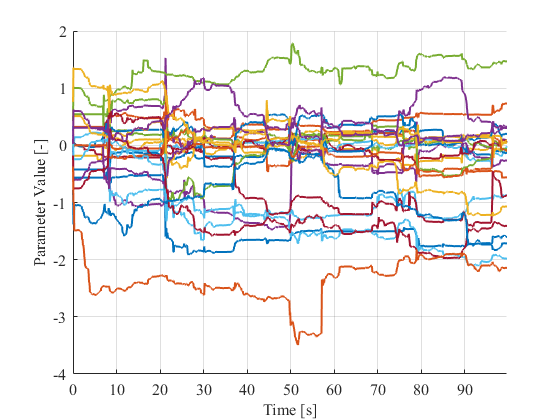}
        \caption{}
        \label{subfig:param_irregular}
    \end{subfigure}


    \begin{subfigure}[b]{0.48\textwidth}
        \includegraphics[width=\textwidth]{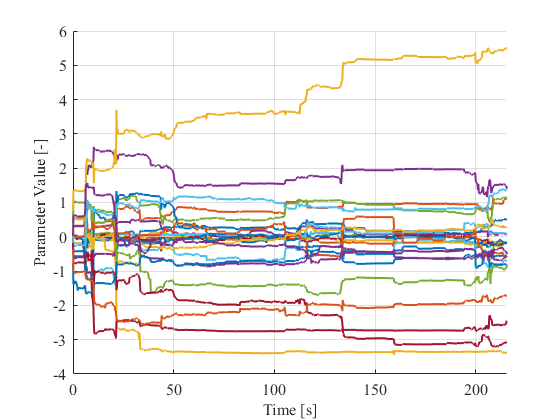}
        \caption{}
        \label{subfig:param_lawnmower}
    \end{subfigure}
    \hfill 
    \begin{subfigure}[b]{0.48\textwidth}
        \includegraphics[width=\textwidth]{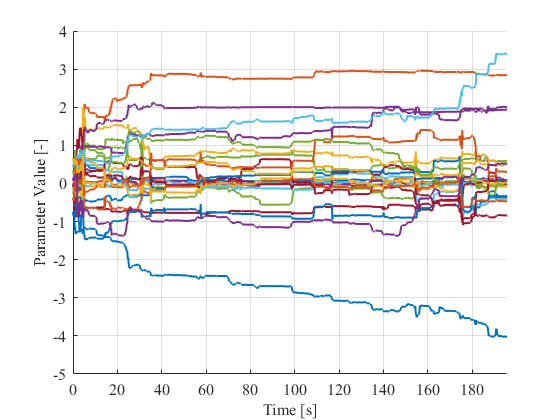}
        \caption{}
        \label{subfig:param_spiral}
    \end{subfigure}
    
    \caption{EKF-based online adaptation of the NN parameters $\bm{x}_{\text{NN}}$ for $\bm{\phi} = \bm{\phi}_{\text{m}}$, $N_h = 8$ for the (a) Figure Eight, (b) Irregular, (c) Lawnmower, and (d) Spiral trajectories.}
    \label{fig:param_convergence}
\end{figure}

\cref{fig:traj_comparison} illustrates the filter's primary function. The raw odometry trajectories diverge significantly from the ground truth, whereas the EKF-NN estimate remains tightly aligned to it, successfully bounding the drift.
This high accuracy, despite a strong unknown, nonlinear interference signal on the magnetometer signal, is enabled by the online learning of the interference model, as shown in \cref{fig:lawnmower_performance} for the lawnmower-trajectory. Increasing the number of neurons leads to a better model fit.
The \ac{NN}'s output closely tracks the nonlinear pattern of the true platform interference field. The learning process itself is visualized in \cref{fig:param_convergence}, showing the network's parameters adapting dynamically as the filter processes measurements. Their continuous adjustment is the expected behavior as the network learns a function of the constantly changing inputs along the trajectory.
Collectively, these results demonstrate the fundamental capability of the \ac{EKF}-\ac{NN} framework. Even a small network integrated into the filter can effectively learn and compensate for complex, unmodeled interference on the measurement signal online, and improve state estimation accuracy.

\subsection{Parallels to Airborne Magnetic Anomaly Navigation}
\label{ssec:parallels}

This simplified model is a direct analog to airborne \ac{MagNav}. While \ac{MagNav} employs an \ac{ESKF} and a full inertial navigation state, the core strategy of simultaneous estimation remains identical. The measurement model encompasses, in addition to the external magnetic field $B_{\text{ext}}$ (anomaly, core, and space weather), a hybrid model of the platform's magnetic interference field that combines the physics-based \ac{TL} model ($f_{\text{TL}}$) for dominant interference and a \ac{NN} ($\tilde{g}_{\text{res}}$) for residual nonlinear interference:
\begin{equation}
    h(\hat{\bfx}_{\text{aug}, t|t-1}) = B_{\text{ext}}(\hat{\bfx}_{t|t-1}) + f_{\text{TL}}(\hat{\boldsymbol{\psi}}_{t|t-1}, \dot{\bfm}_t, m_t, \hbobeta_{\text{TL}, t|t-1}) + \tilde{g}_{\text{res}}(\bophi_{\text{i},t}, \hat{\boLambda}_{t|t-1}).
    \label{eq:measurement_eq_hybrid_model}
\end{equation}
$\boldsymbol{\psi}$ is the heading derived from vector magnetometer $\bfm$, and $m$ is the scalar measurement. The augmented state vector contains navigation errors, \ac{TL} parameters $\boldsymbol{\beta}_{\text{TL}}$, and \ac{NN} weights $\boLambda_{\text{NN}}$, allowing the \ac{EKF} update to simultaneously correct the trajectory and learn the parameters of both models online.

%% file: filter_design.tex
\section{NN-augmented EKF for Online Calibration in MagNav}
\label{sec:filter_design}
This section details the error-state \ac{EKF} fusing inertial navigation data with magnetic anomaly measurements, incorporating an online calibration of platform magnetic interference using both the \ac{TL} model and a neural network.
\paragraph*{System Model}
The error-state dynamics for $\delta \bm p$, $\delta \bm v$, and  $\delta \boepsilon$ are derived by linearizing the global inertial navigation equations \cite{Titterton}, resulting in the so-called Pinson error state model. We augment this model to the $(40+N_p)$-dimensional state vector:
\begin{equation}
    \bfx = \begin{bmatrix}
        \delta \bfp & \delta \bfv & \delta \boepsilon & \delta h_b & \delta a_h & \bfa_b & \boomega_b & S_{TV} & S_{CB} & \bobeta_{TL} & \bm \Lambda_{\text{NN}} & \bfm
    \end{bmatrix}^\top,
\label{eq:state}    
\end{equation}
where $\delta \bfp = \begin{bmatrix}
    \delta \mu & \delta \lambda & \delta \xi
\end{bmatrix}$ are the position errors in geographical coordinates (latitude, longitude, and altitude), $\delta \bfv = \begin{bmatrix}
    \delta v_n & \delta v_e & \delta v_d
\end{bmatrix}$ are the NED-velocity errors, and $\delta \boepsilon = \begin{bmatrix}
    \delta \phi & \delta \theta & \delta \psi
\end{bmatrix}$ are the attitude errors in body-frame.

$\delta h_b$ and $\delta a_h$ are the aiding altitude error and vertical acceleration error, respectively. As in \cite{gnadt_advanced_2022}, no physical barometer measurement is used. Instead, a third-order dynamic control loop within the state transition matrix modifies the unstable INS vertical channel errors ($\delta h$, $\delta v_d$) to the stable, bounded dynamics of the $\delta h_b$ state to mathematically prevent divergence of the vertical states.
The states $\bfa_{\bfb}$ and $\boomega_{\bfb}$ are modeling the accelerometer and gyroscope biases as random walks. 
The bias estimates don't correct the IMU measurements directly but are integrated separately and directly provide corrections to the position, velocity, and attitude error estimates. 
$S_{TV}$ is modeling a random walk tuned to the typical characteristic standard deviation and timescale of space-weather disturbances on the total magnetic field. $S_{CB}$ models a constant measurement bias, i.e., a magnetic map offset or sensor calibration bias.
$\bfx_{TL}$ and $\bm \Lambda_{\text{NN}}$ are the parameters of the platform calibration models. $\bfx_{TL}$ are the 18 parameters of the \ac{TL} model, $\bm \Lambda_{\text{NN}}$ is the vector with the $N_p$ \ac{NN} parameters. This filter is similar to the Online \ac{TL} Calibration Filter for MagNav \cite{Canciani2022}, extended by the barometer loop from Gnadt \cite{gnadt_advanced_2022} and the neural network parameters.
\paragraph*{State Dynamics}
The dynamics of the continuous-time system are described by
\begin{align}
\dbfx = \bfF \bfx + \bfG \bfu + \bfw
\end{align}
where $\bfF$ describes dynamic evolution of the total system. The vector magnetometer measurements are treated as a pseudo-control input $\bfu = \bfm(t)$ driving the states $\bfm$ via a first-order lag with a very small time constant $\tau \to 0$. This models the instantaneous replacement of the state with the measurement. The input mapping matrix is
$\bfG = [
        \bfzero_{3 \times (37+N_p)} ,\, \frac{1}{\tau} \bfI_{3 \times 3}
    ]^\top$, which ensures that the discrete-time equivalent input matrix approaches identity, $\bfG \approx \bfI$, as $\tau \to 0$.

The full continuous-time system matrix, expressed in block matrices according to the state vector in \eqref{eq:state}, is 
\begin{align} \bfF = \text{diag}(\bfF_{\text{NAV}}, \bfF_{\text{IMU}}, \bfF_{\text{MAG}}, \bfF_{\text{TL}}, \bfF_{\text{NN}}, \bfF_{\text{VMAG}})_{(40+N_p) \times (40+N_p)}
\end{align} 
$\bfF_{NAV}$ contains the terms for the navigation and barometer states, $\bfF_{\text{IMU}}$ the accelerometer and gyro bias terms, 

\begin{align} 
    \bfF_{\text{NAV}} &= \begin{bmatrix} 
                    \bfF_{pp} & \bfF_{pv} & \bfzero_{3 \times 3} & \bfF_{pb} \\
                    \bfF_{vp} & \bfF_{vv} & \bfF_{v\epsilon} & \bfF_{vb} \\ 
                    \bfF_{\epsilon p} & \bfF_{\epsilon v} & \bfF_{\epsilon \epsilon} & \bfzero_{3 \times 2} \\
                    \bfF_{bp} & \bfzero_{2 \times 3} & \bfzero_{2 \times 3} & \bfF_{bb}
                  \end{bmatrix}_{11 \times 11} \quad
\bfF_{\text{IMU}} = \text{diag}(\bfF_{aa}, \bfF_{\omega \omega})_{6 \times 6}, 
\end{align}
and the off-diagonal blocks linking the navigation states to the IMU biases are: 
\begin{equation}
    \bfF[4:6, 12:14] = \bfF_{va},\quad
    \bfF[7:9, 12:14] = \bfF_{\epsilon a},  \quad
    \bfF[7:9, 15:17] = \bfF_{\epsilon \omega} 
\end{equation}
where, for readability, the $\delta$ was dropped in the subscripts of block-matrices relating to error-state dynamics. The Pinson error model blocks ($\bfF_{pp}, \bfF_{pv}$, etc.) may be found in \cite{Titterton} or \cite{canciani_absolute_2016}. The blocks for the third-order barometer loop are given as in \cite{gnadt_advanced_2022}:
\begin{align}
\bfF_{pb} = \begin{bmatrix} 0 & 0 \\ 0 & 0 \\ k_1 & 0 \end{bmatrix}, \quad
\bfF_{vb} = \begin{bmatrix} 0 & 0 \\ 0 & 0 \\ -k_2 & 1 \end{bmatrix}, \quad
\bfF_{bp} = \begin{bmatrix} 0 & 0 & 0 \\ 0 & 0 & k_3 \end{bmatrix}, \quad
\bfF_{bb} = \begin{bmatrix} -1/\tau_b & 0 \\ -k_3 & 0 \end{bmatrix}
\end{align}
The dynamics matrix of $S_{TV}$ and $S_{CB}$ is $\bfF_{\text{MAG}} = \mathrm{diag} \begin{bmatrix}
-1/\tau_{TV} & 0 \end{bmatrix}$, modeling fast-varying temporal variations with a correlation time $\tau_{TV}$, and a constant, zero-noise bias state.
The entries of the transition matrix for both the \ac{TL} Model and the \ac{NN} are set to zero, $\bfF_{\text{TL}} = \bfzero_{18 \times 18}$ and $\bfF_{\text{NN}}  = \bfzero_{N_p \times N_p}$. By using this static model, we assume the model parameters to vary slowly, driven by the noise. The slowly varying behavior is captured by the process noise.
The diagonal entries of the last block-matrix $\bfF_{\text{VMAG}}$ are set to a large negative number ($-\frac{1}{\tau} \approx -\infty$) such that the discretized dynamics matrix contains entries $ e^{- \Delta t/\tau} \approx 0$, to zero the vector magnetometer states from the previous time step and subsequently overwrite them by the new measurements provided as control input $\bfu$.

The continuous process noise covariance matrix $\covQ_c$ is a block diagonal matrix
\begin{align}
\covQ_c = \text{diag}\left(\bfzero_3, \sigma_{\bfa}^2 \bfI_3, \sigma_{\boomega}^2 \bfI_3, \frac{2\sigma_b^2}{\tau_b}, 0, \frac{2 \sigma_{\bfa \bfb}^2}{\tau_{\bfa \bfb}} \bfI_3, \frac{2 \sigma_{\boomega \bfb}^2}{\tau_{\boomega \bfb}} \bfI_3, \frac{2\sigma_{tv}^2}{\tau_{tv}}, \sigma_{CB}^2, \sigma_{TL}^2 \bfI_{18}, \sigma_{\text{NN}}^2 \bfI_{N_p}, \sigma_m^2 \bfI_3\right)
\end{align}
The random walk effect of the noise on the velocity and attitude states is modeled by $\sigma_{\bfa}^2$ (Velocity Random Walk) and $\sigma_{\boomega}^2$ (Angular Random Walk). The biases for the barometer, \ac{IMU}, and $S_{\text{TV}}$ are modeled as \ac{FOGM} processes driven by noise. The calibration parameters ($S_{\text{CB}}, \bfx_{\text{TL}}, \bm \Lambda_{\text{NN}}$) and the vector magnetometer state ($\bfm$) are modeled as random walks, meaning their variance grows linearly with the number of time steps, $\sum_{t=0}^{N} \bfw_{t,i} \sim \mathcal N(0, N \bosigma_i^2)$.
The continuous-time system is discretized over the sampling interval $\Delta t = t_{k+1}-t_k$. Assuming that $\bfF$ is constant over this interval, we compute the transition matrix $\boPhi_{k+1, k} \approx e^{\bfF \Delta t}$ and the discrete process noise covariance $\covQ_d \approx \covQ_c \Delta t$. The error state mean and covariance are then predicted using the standard \ac{EKF} time-update equations.

\paragraph*{Error-state update}
The state vector \cref{eq:state} is updated using \cref{eq:ekf_update} for the magnetic field measurements.
The measurement model is
\begin{align}
\begin{split}
       z_t &= \bfh(\hbfx_{t|t-1}) + \upsilon_t  \\
       &= B_\text{anom}(\hbfx_{t|t-1}) + B_\text{core}(\hbfx_{t|t-1}) + B_\text{bias}(\hbfx_{t|t-1},t)  + B_\text{pf}(\hbfx_{t|t-1}, t) + \upsilon_t, \quad \upsilon \sim \mathcal{N}(0, \sigma_{\upsilon}^2) \\ 
\end{split}
\end{align}
The first two terms are the two main components of the Earth's magnetic field: The magnetic anomaly field $B_\text{anom}(\hbfx)$, which is our final navigation signal represented on a magnetic anomaly map, and the contribution of the Earth's core field, $B_\text{core}(\hbfx)$, modeled by the \ac{IGRF} global magnetic field model.
The term $B_\text{bias}(\hbfx_{t|t-1}, t)$ models space-weather-induced fast temporal variation in the magnetic field and slowly varying biases (including map errors) by adding up the two bias states
\begin{align}
    B_\text{bias}(\hbfx_{t|t-1}) = S_{TV,t|t-1} + S_{CB,t|t-1}.
\end{align}
The platform field $B_\text{pf}(\hbfx_{t|t-1})$ is given by the hybrid interference model represented by a linear combination of the \ac{TL} model and a \ac{NN}, as shown in eq. \cref{eq:measurement_eq_hybrid_model}:
\begin{alignat}{2}
    B_\text{pf}(\hbfx_{t|t-1}) &= f_{\text{TL}}(\bfm_t, \dot{\bfm_t}, m_t, \hbobeta_{\text{TL}, t|t-1}) &&+ \tilde{g}_{\text{res}}(\bophi_t, \hat{\boLambda}_{t|t-1}) \\
    &= \bfA_{\text{TL}}(\bfm_t, \dot{\bfm_t}, m_t) \hbfx_{TL, t|t-1} &&+ B_{\text{NN}}(\bophi_t, \hat{\boLambda}_{t|t-1}).
    \label{eq:noise_model_combined}
\end{alignat}
Both the \ac{TL} parameters $\bobeta_{\text{TL}}$ and the network parameters $\boLambda$ are learned online during navigation by the \ac{KF}.
A subtlety here is that we can recognize the neural network model's adaptive nature by noticing its implicit dependency on the previous time step via $\boLambda_{t|t-1}$. This renders feature engineering, like using time derivatives of the magnetometer measurements as network input, unnecessary.
The \ac{TL} model uses the vector magnetometer measurements $\bfm$ as a representation for attitude \cite{Gnadt2022a}. 
The input feature vector $\bophi$ for the network may generally contain any state variable or sensor measurement available in the dataset, but we restrict ourselves to magnetometer measurements only.
Both interference models receive feedback from the current magnetometer measurements ($\bfm$ is the vector, $m$ the scalar magnetometer measurement) and get updated simultaneously to the INS states by the filter. 
The \ac{TL} $\bfA$-matrix contains time derivatives of vector magnetometer measurements that are calculated as finite differences $\dot{\bfm}_t \approx (\bfm_t - \bfm_{t-1})/dt$.

\paragraph{Jacobian}
Since the parameters of our platform field calibration model consist of both the \ac{TL} model and a Neural Network, the Jacobian $\bfH$ is 
\begin{align}
 \bfH = \begin{bmatrix}
    \frac{\partial \bfh}{\partial lat} & \frac{\partial \bfh}{\partial lon} & 0 & \bfzero_{1 \times 3}
    & \bfzero_{1 \times 3} & \bfzero_{1 \times 2} & \bfzero_{1 \times 3} & \bfzero_{1 \times 3} & 1 & 1 & \bfA_{\text{TL}} & \frac{\partial \bfh}{\partial \boLambda_{\text{NN}}} & \frac{\partial \bfh}{\partial \bfm}
    \end{bmatrix},
\end{align}
The derivatives with respect to the horizontal position are calculated as finite differences from the three-dimensional core field and map function.
The derivatives with respect to the \ac{TL} parameters yield $\bfA_{\text{TL}}$, as can be read off directly from eq. \cref{eq:noise_model_combined}. The gradient of the network with respect to its parameters, $\frac{\partial \bfh}{\partial \boLambda_{\text{NN}}}$ is calculated using the chain rule on the nested function as described in Appendix~\ref{sec:nn_jacobian}. The derivatives $\frac{\partial \bfh}{\partial \bfm}$ are composed of the partial derivatives of the \ac{TL} model with respect to the vector magnetometer measurement states and the terms $\frac{\partial \tilde{\bfg}_{res}}{\partial \bfm}$ (see Appendix~\ref{ssec:jacobian_mvec} for the full expression).

\paragraph*{Correction of the Nominal State}
We recall that our filter state is partly composed of error states. 
The corrected position estimate $\hbfp$ is calculated as the composition of the drifting INS solution $\bfp_{\text{ins}}$ (as obtained by integrating accelerometer and gyroscope raw measurements) with the growing position errors $\delta \hbfp$, $\hbfp = \bfp_{\text{ins}} + \delta \hbfp$.

%% file: implementation.tex
\section{Implementation}
\label{sec:implementation}
\subsection{Dataset and Methodology}

\begin{figure}[tbh!]
    \centering
    \includegraphics{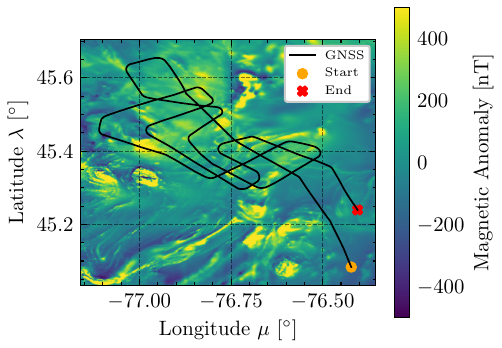}
    \caption{GNSS trajectory of flight line 1007.06.}
    \label{fig:gnss_trajectory}
\end{figure}

For our primary analysis, we selected flight line 1007.06 (see \cref{fig:gnss_trajectory}) from the MagNav challenge dataset \cite{gnadt_daf-mit_2023} for comparability with prior work \cite{gnadt_advanced_2022}. Results for the flight line 1003 can be found in Appendix~\ref{ssec:flight1003_results}.
We consider two initialization scenarios that differ in the amount of prior information provided to the model:
\begin{enumerate}
    \item \textbf{Cold Start (C):} The filter is initialized with uninformative priors. The \ac{TL} coefficients, map bias, and \ac{NN} parameters start at zero (or random initialization for \ac{NN} weights), and the state covariance matrix is initialized to identity (scaled by the gain factor), reflecting no prior knowledge.
    \item \textbf{Warm Start (W):} The final calibration model parameter estimates ($\hbfx_{\text{TL}}$, $\hat{S}_{\text{CB}}$ and $\hat{\beta}_{\text{TL}}$) and their state covariance matrices from a cold start run of the same trajectory are used to initialize the filter, mimicking memory of a prior flight.
\end{enumerate}
Our neural network architecture consists of a small single hidden layer with a $\mathrm{tanh}(\cdot)$ activation, followed by a single-neuron output layer without a bias term or activation function. This choice of a shallow network with low neuron numbers is supported by our results in \cref{sec:toymodel} and \cite{gnadt_advanced_2022}. The output bias is omitted to avoid coupling with the constant map bias $S_{\text{CB}}$, which would otherwise share the same Jacobian. All inputs are normalized, and the network output is denormalized using a scaling factor of $\alpha = \SI{400}{\nano\tesla}$, and no offset, to match the expected range of residual platform interference after \ac{TL} compensation. We initialize the weights according to the Glorot method \cite{glorot_understanding_2010}, where the initial weights and biases for a layer $L$ are defined by:
\begin{align}
\bfW_{ij} \sim \mathcal{N}\left(0, \sigma_L^2 \right)
\quad \text{where} \quad
\sigma_L = \gamma \sqrt{\frac{2}{\text{fan}_{\text{in}} + \text{fan}_{\text{out}}}},
\qquad
\bfb_i = 0,
\end{align}
with a gain factor $\gamma = 10^{-2}$. $\text{fan}_{\text{in}}$ and $\text{fan}_{\text{out}}$ denote the number of input and output units of the layer $L$. The initial state covariance and process noise covariance for the neural network parameters for the cold-start are chosen as $\covP_{0,\mathrm{NN}} = \bfI$ and $\covQ_{\mathrm{NN}} = 10^{-20}\bfI$.
The covariance matrix $\covP_t$ functions as an adaptive step-size matrix (or preconditioning matrix) for the online parameter updates \cite{Ollivier2018}. By setting $\covP_{0,\mathrm{NN}}$ to the identity matrix, we effectively impose an isotropic initial learning rate to all parameters. 
The measurement noise covariance is fixed to $\covR = \SI{10}{\nano\tesla\squared}$. These design choices are consistent with the EKF--online natural gradient interpretation outlined in \cref{ssec:tuning_theory}, where only the initial covariance $\covP_0$ and the ratio $\covQ/\covR$ need to be tuned to control the effective learning rate. The relatively large $\covP_{0,\mathrm{NN}}$ compared to $\sigma_L^2$ acts as a strong prior on the \ac{NN} parameters, resulting in a large initial learning rate during the transient phase. Furthermore, by fixing $\covR$ and selecting a strictly positive but very small process noise covariance $\covQ_{\mathrm{NN}}$, the EKF realizes a non-decaying yet low steady-state learning rate, so that new measurements are incorporated into the \ac{NN} parameters in a conservative, stable manner as they enter the filter. Assuming no prior knowledge of the \ac{TL} calibration coefficients and the constant measurement bias $S_{\text{CB}}$ is assumed for the cold start, they are initialized as $\bobeta_{\mathrm{TL},0} = \bfzero_{18\times 1}$ and $S_{\text{CB},0} = 0$.
The initial state covariance and the process-noise covariance matrices for the \ac{TL} parameters are set to $\covP_{0,\text{TL}}=10^{5}\bfI$ and $\covQ_{\text{TL}}=\bfI$.
For the warm start, the final state estimates of the \ac{NN} and \ac{TL} parameters and their covariance matrices $\covP_{\text{end}}$ from the cold start are reused (see \cref{fig:nn_sigma} as an example).
The model was provided with a magnetometer-only feature set consisting of data from the Flux A sensor and the corresponding uncompensated magnetometers 1, 2, 3, 4, and 5, which were also used for map-matching. The notoriously noisy Magnetometer 2 that exhibited artifacts when compensated using traditional band-pass-filter-based \ac{TL} calibration flights, is included here. Magnetometer 1 ($\approx$ ground truth) and the low-noise Magnetometer 5 are included to establish a baseline performance.
To improve robustness, an innovation-based outlier rejection scheme is applied: measurements whose normalized innovation squared exceeds a $\chi^2$ threshold of $6$ are excluded from the update after the first \SI{10}{\minute} of flight.
Performance is evaluated using the horizontal position \ac{RMSE}, or \ac{DRMS}, as it is often referred to in the navigation domain, to quantify the positioning error,
\begin{equation}
\mathrm{DRMS}
= \sqrt{\frac{1}{N}\sum_{t=1}^{N}
\left[
(x_t^{\mathrm{GNSS}} - x_t^{\mathrm{MagNav}})^2
+
(y_t^{\mathrm{GNSS}} - y_t^{\mathrm{MagNav}})^2
\right] },
\end{equation}
where $x_t^{\mathrm{GNSS}}, y_t^{\mathrm{GNSS}}$ and $x_t^{\mathrm{MagNav}}, y_t^{\mathrm{MagNav}}$ are the \ac{GNSS} and \ac{MagNav} positions expressed as \ac{UTM} eastings and northings in meters. The error of the aircraft magnetic field prediction is quantified by 
\begin{equation}
\mathrm{RMSE}_{m}
= \sqrt{\frac{1}{N}\sum_{t=1}^{N}
\left(m_t - h(\bfx_t)\right)^2 },
\end{equation}
that implicitly evaluates how well the calibration model predicts the platform magnetic field.

\subsection{Results}
\begin{figure}[tb]
    \centering

    \includegraphics[width=\linewidth]{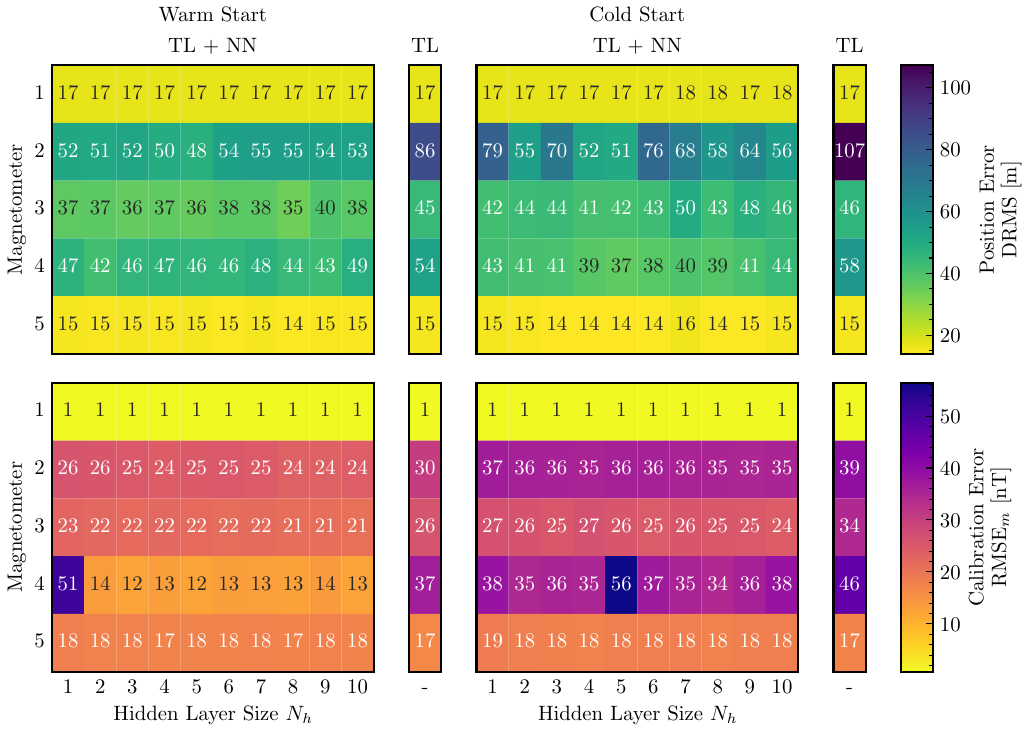}
    
    \caption{Magnetic aiding performance results using magnetometer 1 to 5 and different hidden layer sizes $N_h$.}
    \label{fig:drms_rmse_heatmaps_tanh}
\end{figure}

\begin{figure}[tb]
    \centering
    \begin{subfigure}[t]{0.48\linewidth}
        \centering
        \includegraphics[width=\linewidth]{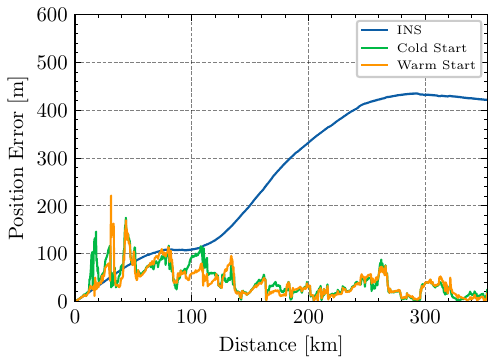}
        \caption{}
        \label{fig:cold_warm_nn_tl}
    \end{subfigure}
    \hfill
    \begin{subfigure}[t]{0.48\linewidth}
        \centering
        \includegraphics[width=\linewidth]{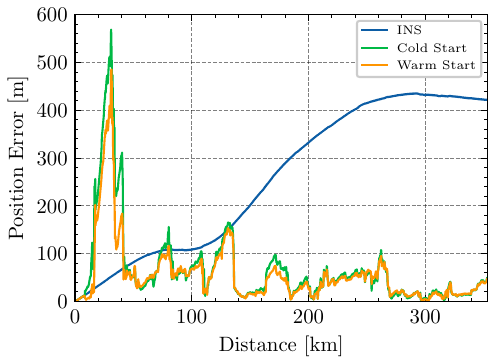}
        \caption{}
        \label{fig:cold_warm_tl}
    \end{subfigure}

    \caption{Positioning accuracy in cold and warm start scenarios using Magnetometer 2 ($N_h = 5$). (a) Our hybrid approach combining \ac{TL} and \ac{NN} calibration (\ac{DRMS} cold start \SI{51}{\meter} / warm start \SI{48}{\meter}) (b) \ac{TL} online calibration ($N_h = 0$), \ac{DRMS} cold start \SI{107}{\meter} / warm start \SI{86}{\meter}).}
    \label{fig:cold_warm_comparison}
\end{figure}

\cref{fig:drms_rmse_heatmaps_tanh} illustrates the position and calibration accuracies achieved during the \SI{87}{\minute} flight. While the unaided navigation-grade INS solution drifted to an endpoint error of \SI{422}{\meter}, our algorithm binds this drift consistently using any of the three noisiest in-cabin magnetometers of the dataset, regardless of the initialization method (cold  or warm start). We tested the algorithm for different numbers of neurons in the hidden layer, $N_h$, against the \ac{TL} online calibration only case where no \ac{NN} is used for calibration. 
The positioning accuracy is robust and clearly improved by factors of $\approx 10-90 \%$ over the \ac{TL}-only case across all magnetometers that are exposed to platform interference, though the warm start provides a slight advantage for the high-interference Magnetometer 2 (averaged uncompensated interference $\approx \SI{1250}{\nano\tesla}$) and Magnetometer 3 (averaged uncompensated interference $\approx \SI{1150}{\nano\tesla}$). In contrast, prior information proves less significant for the lower-interference Magnetometer 4 (averaged uncompensated interference $\approx$ \SI{250}{\nano\tesla}) and Magnetometer 5 (averaged uncompensated interference $\approx$ \SI{100}{\nano\tesla}). Regarding the hidden layer size a slight trend is revealed: the calibration RMSE tends to decrease with increasing network size for most cases. However, the resulting effect on position error is more ambiguous. For any $N_h > 1$, no consistent trend across all magnetometers is visible. 
We compare these results against the ``Online Model 2c'' approach from \cite{gnadt_advanced_2022}, which achieved \ac{DRMS} errors of \SI{32}{\meter} (Mag 3), \SI{37}{\meter} (Mag 4) and \SI{18}{\meter} (Mag 5).
Their approach used a similar feature set, including time-derivatives in the network input, and relied on extensive pre-training using a calibration flight and \SI{12}{\hour} of flight data to initialize and train the \ac{TL} coefficients and a \ac{NN} correction term. In comparison, our method achieves similar positioning performance without requiring dedicated calibration flights or initial information based on a large dataset, even for the very noisy Magnetometer 2, demonstrating the efficacy of a more flexible architecture in handling varying magnetic interference levels. Specifically, for $N_h=5$ in the cold-start case we obtain $\mathrm{DRMS}$ values of \SI{42}{\meter} (Mag 3), \SI{37}{\meter} (Mag 4), and \SI{14}{\meter} (Mag 5).
To explore the calibration and navigation behavior more closely, we take a detailed look at the algorithm's behavior for the noisy Magnetometer 2 (detailed analyses for magnetometers 1-5 can be found in Appendix~\ref{ssec:mag1_suppl}--\ref{ssec:mag5_suppl}). In \cref{fig:cold_warm_comparison}, the time evolution of the position errors for our hybrid architecture in the cold- and warm-start scenarios (\cref{fig:cold_warm_nn_tl}) are compared with the position errors running online calibration only using the \ac{TL} model (\cref{fig:cold_warm_tl}, similar to \cite{Canciani2022}, but initialized without prior knowledge).
The better position accuracy of our hybrid model is particularly significant during the first \SI{50}{\kilo\meter} of the flight, a behavior that could be particularly advantageous for low-grade \ac{INS} systems that drift quickly.
\begin{figure}[tb]
    \centering
    \begin{subfigure}[t]{0.48\linewidth}
        \centering
        \includegraphics[width=\linewidth]{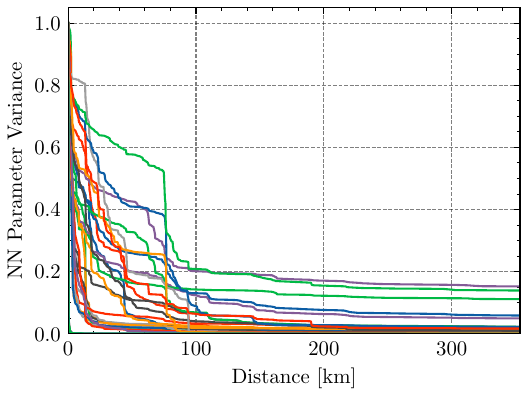}
        \caption{}
        \label{fig:nn_sigma_cold_start}
    \end{subfigure}
    \hfill
    \begin{subfigure}[t]{0.48\linewidth}
        \centering
        \includegraphics[width=\linewidth]{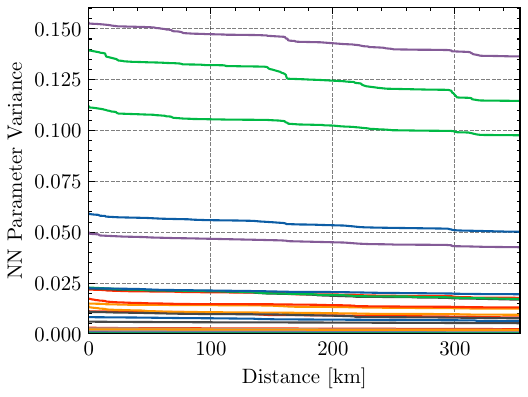}
        \caption{}
        \label{fig:nn_sigma_warm_start}
    \end{subfigure}

   \caption{Evolution of the \ac{NN} parameter variance $\covP_{\text{NN}}$ using Magnetometer 2 ($N_h = 5$), comparing (a) cold start and (b) warm start.}
   \label{fig:nn_sigma}
\end{figure}

\begin{figure}[tb]
    \centering
    \begin{subfigure}[t]{0.45\linewidth}
        \centering
        \includegraphics[width=\linewidth]{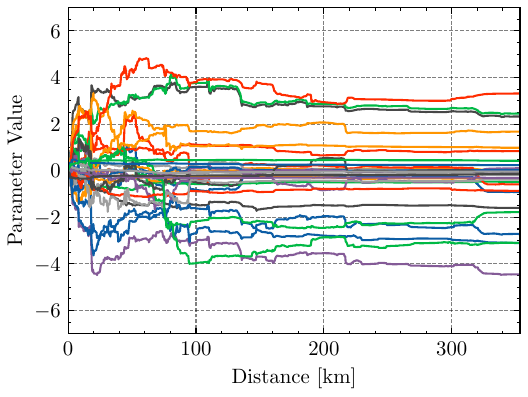}
        \caption{}
        \label{fig:nn_cold_start}
    \end{subfigure}
    \hfill
    \begin{subfigure}[t]{0.45\linewidth}
        \centering
        \includegraphics[width=\linewidth]{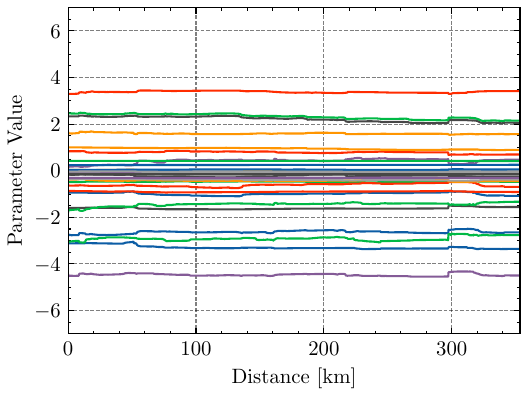}
        \caption{}
        \label{fig:nn_warm_start}
    \end{subfigure}

    \hfill
    \begin{subfigure}[t]{0.48\linewidth}
        \centering
        \includegraphics[width=\linewidth]{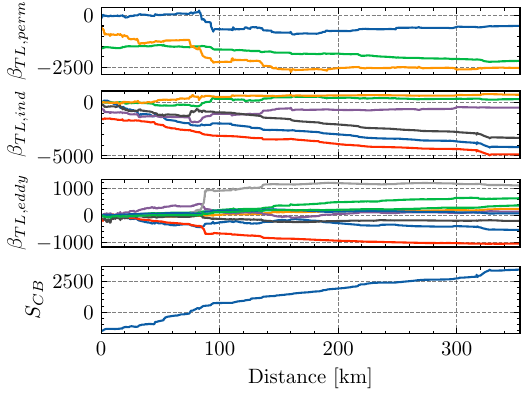}
        \caption{}
        \label{fig:tl_parameters_cold}
    \end{subfigure}
    \hfill
    \begin{subfigure}[t]{0.48\linewidth}
        \centering
        \includegraphics[width=\linewidth]{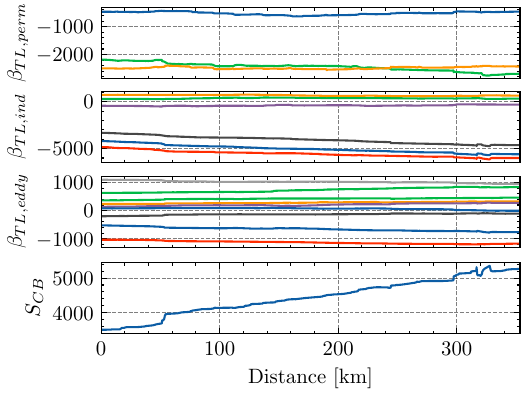}
        \caption{}
        \label{fig:tl_parameters_warm}
    \end{subfigure}

    \caption{Evolution of \ac{NN} (top) and \ac{TL} (bottom) model parameters using Magnetometer 2 ($N_h = 5$), comparing (a), (c) cold start and (b), (d) warm start.}
    \label{fig:tl_parameters}
\end{figure}

The evolution of the \ac{NN} parameters is shown in \cref{fig:nn_cold_start,fig:nn_warm_start}. When `cold-started' (\cref{fig:nn_cold_start}), they begin at their near zero initialization values and get adapted strongly during the first 50 to \SI{80}{\kilo\meter}, settling eventually. This indicates that the \ac{NN} has largely converged and only undergoes slow updates driven by the small steady-state learning rate $\covP_{\text{NN}}$ defined by $\covQ/\covR$. When initialized with the memory from a previous flight (\cref{fig:nn_warm_start}), the parameters vary only slightly over the entire trajectory since the \ac{EKF} only has to fine-tune rather than largely correct them. The evolution of the variance of the \ac{NN} parameters on the diagonal of $\bfP_{\text{NN}}$ in \cref{fig:nn_sigma} shows how the learning rates converges from initially large values in the cold start to allow for learning, in contrast to the smaller, already converged system of the warm start.
The complementary evolution of the \ac{TL} parameters in \cref{fig:tl_parameters_cold} and \cref{fig:tl_parameters_warm} shows a similar behavior.
Only the constant map-bias term $S_{CB}$ exhibits a mostly monotonic increase from zero to its final value, indicating that the EKF attributes a substantial portion of the uncompensated field to a large constant offset. 

\begin{figure}[tb]
    \centering
    \begin{subfigure}[t]{0.48\linewidth}
        \centering

        \includegraphics[width=\linewidth]{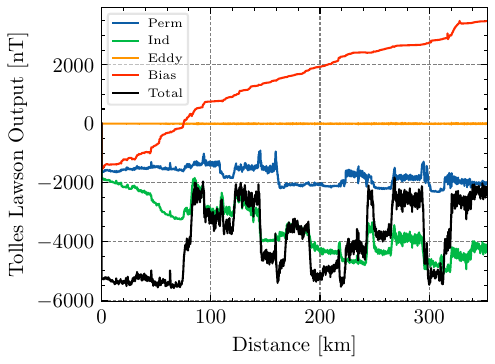}
        \caption{}
        \label{fig:tl_components}
    \end{subfigure}
    \hfill
    \begin{subfigure}[t]{0.48\linewidth}
        \centering
        \includegraphics[width=\linewidth]{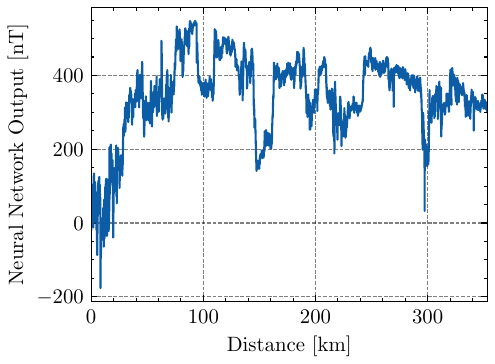}
        \caption{}
        \label{fig:nn_output}
    \end{subfigure}

    \caption{Comparison between the (a) \ac{TL} model compensation components (including the bias state $S_{CB}$) and the (b) learned \ac{NN} output using Magnetometer 2 in the cold start scenario.}
    \label{fig:tl_vs_nn_output}
\end{figure}

The resulting calibration model components for the cold-start scenario from these two models, the deterministic \ac{TL}-model and the corrective \ac{NN} part, are shown in \cref{fig:tl_vs_nn_output}.
\cref{fig:tl_components} illustrates that the \ac{TL} model captures the majority of the platform interference at an order of magnitude of thousands of nanoteslas, accounting for the dominant permanent and induced field components. The constant bias term $S_{\text{CB}}$ increases continuously, showing a behaviour seemingly anti-correlated to the permanent and induced compensation components. The total \ac{TL} output (black curve) exhibits characteristic step-like changes that follow the aircraft maneuvers. The NN output (\cref{fig:nn_output}) settles at the order of a few hundred nanoteslas, consistent with the chosen output scaling factor of \SI{400}{\nano\tesla}. This confirms that the \ac{NN} operates as a residual learner, modeling only the higher-frequency nonlinearities left uncorrected by the physical model. By bounding the \ac{NN} to a fine-scale corrective role by keeping its learning rate small, we can mitigate the risk of the network overfitting to large biases or introducing instability into the magnetic calibration and navigation.

\subsection{Discussion}
Our results demonstrate that the proposed hybrid architecture successfully connects purely model-based calibration and data-driven model learning. Although yielding comparable or slightly higher \ac{DRMS} values on the MagNav benchmark dataset than methods requiring extensive pre-training \cite{gnadt_machine_2022}, our approach prioritizes operational flexibility. By enabling effective performance in both cold-start scenarios and warm starts with minimal initialization data, we eliminate the need for dedicated calibration flights or large historical datasets. We also avoid feature engineering or integration with multiple auxiliary aircraft systems. This gain in operational readiness outweighs the slight trade-off in absolute accuracy.
An important operational consideration is the 50-\SI{100}{\kilo\meter} convergence period observed during ``cold starts''. 
During this phase, the system’s integrity can be monitored via the augmented state covariance matrix $P$. 
Convergence is identified when the NN parameter variances $\covP_{\text{NN}}$ and their cross-covariances with navigation states settle into a steady-state regime defined by the $\covQ/\covR$ ratio. Furthermore, as the model matures, the magnetic innovation should approach zero-mean white noise, indicating that the filter has successfully decoupled aircraft interference from the navigation signal.
Practically, the initial \ac{MagNav} solution can be cross-checked with a navigation-grade INS or GNSS-aided navigation solution if available.
The decoupled architecture acts as a safety enabler for aviation. The physics-based \ac{TL} model captures bulk interference (thousands of \si{\nano\tesla}), whereas the \ac{NN} is architecturally constrained to a residual learning role via a \SI{400}{\nano\tesla} output scaling factor. Though introducing real-time adaptation of a network may introduce the problem of verifying a non-static system, this separation mitigates potential ``black box'' instability by anchoring the system in physics while confining the adaptive element to modeling local nonlinearities. Regarding network topology, we observe that increasing the hidden layer size $N_h$ improves calibration \ac{RMSE} slightly but yields ambiguous positioning results. We hypothesize that larger networks reduce state observability by overfitting to the anomaly map signal or INS drift. Consequently, a minimalist network ($N_h \approx 2-5$) is preferred to prevent the calibration model from cannibalizing the navigation signal, aligning with findings in \cite{gnadt_advanced_2022}.
Unlike standard training often reliant on trial-and-error, the filter parameters possess direct functional interpretations: the initial state covariance $\covP_0$ dictates the initial learning rate (prior uncertainty), whereas the steady-state ratio of $\covQ/\covR$ governs the asymptotic plasticity (or effective memory). This rigorous grounding explains the observed behavior of rapid convergence within the first $\approx \SI{50}{\kilo\meter}$ followed by naturally settling learning rate and confirms the filter's ability to perform online, second-order optimization to recover a magnetic anomaly aided navigation solution from a complete lack of prior knowledge on the platform interference.

\nocite{SciencePlots}

%% file: conclusion.tex
\section{Conclusion}
\label{sec:conclusion}
\vspace{-0.08cm}

This work presented a fully adaptive \ac{EKF} architecture that enables magnetic anomaly navigation while avoiding pre-flight calibration. By augmenting the \ac{KF} state with the parameters of a Neural Network, we achieved simultaneous state estimation and identification of the effect of magnetic platform disturbances. We established that the recursive update is mathematically equivalent to online Natural Gradient descent, providing a state-of-the-art geometry-aware second-order optimization to identify the aircraft’s magnetic signature during flight.
We demonstrated:
\begin{enumerate}
    \item Cold-Start Viability: The system binds INS drift using a combination of MagNav INS aiding and pure online learning, matching the performance of methods that require extensive pre-training. Performance can be improved using information from historical flight data to warm-start the filter.
    \item {Operational Feasibility:} Integrating the Tolles-Lawson model with a Neural Network provides a robust, explainable baseline while capturing complex nonlinear interference. Restricting the Neural Network to a residual learning role preserves the interpretability and mitigates the risk of unconstrained model divergence.
    \item Architectural efficiency: A magnetometer-only feature set and a shallow network architecture are sufficient for navigation-grade performance, facilitating deployment on embedded hardware.
\end{enumerate}

Providing the theoretical foundations for the explainability of this adaptive system, our findings support the operational deployment of magnetic anomaly navigation. 
Future research should address scalability through decoupled or federated architectures like the Partial-Update Schmidt-Kalman Filter, which applies a percentage weight to parameter updates to prevent initial learning transients from destabilizing primary navigation states. Resilience to map artifacts could be improved via Moving Horizon Estimation, while Adaptive Kalman Filtering may optimize the trade-off between fast learning and estimation consistency. Implementing a metric to determine calibration maturity can be useful to enable the system to transition from a provisional status to a high-integrity solution and to facilitate reliable warm starts for subsequent flights. Additionally, investigating tightly-coupled architectures and robust fault detection and exclusion can help to distinguish adaptive model learning from sensor malfunctions.

From a certification perspective, while residual bounding mitigates risk, developing formal convergence proofs remains a challenge. Operational trust can be established by monitoring covariance-based observability metrics and verifying consistency against navigation-grade INS solutions during the initial convergence window . Finally, a rigorous failure mode analysis of the signal chain is essential. In this context, quantum magnetometers may offer benefits in mitigating sensor-level vulnerabilities, further enabling the deployment of high-integrity magnetic mapping and navigation.

%% file: acknowledgements.tex
\section*{Acknowledgment}
The authors would like to thank Mathieu Brunot, Eric Euteneuer and Laurent Azoulai for valuable discussions on airborne data fusion architectures. We also thank Philipp Hartmann for his supervision of the master’s thesis research contributing to this publication.
This work is based in part on ideas studied in \cite{Nebendahl2025}.
The authors acknowledge the use of AI tools (Gemini Pro, ChatGPT Plus, Claude Pro) in the drafting and refinement of this manuscript and in coding the underlying algorithms.

%% file: appendix/tolles_lawson.tex
\section{Tolles-Lawson Calibration}
\label{sec:tolles_lawson}
\ac{TL} calibration encompasses a well-established suite of techniques to calibrate magnetometers onboard aircraft used for geophysical surveys. To avoid interference from aircraft-generated magnetic fields, it is typically tried to place the sensors at magnetically quiet locations. This is often not perfect or possible, and residual interferences from metal parts need to be compensated for in data processing. Additionally, misalignment, sensor errors, and scale factors can further distort the measurements.
It is important to note that the corrections calculated by the \ac{TL} method do, however, not take into account nonlinear transient effects, like turning onboard electric systems on and off or passengers moving devices.
\subsection{The Calibration Model}
The total magnetic vector field measured by an airborne magnetometer can be divided into the Earth's magnetic field $\bfB_e$ and the aircraft magnetic field $\bfB_a$.
\begin{align}
    \bfB_t = \bfB_e + \bfB_a
\end{align}
The aircraft magnetic field can be split into three terms representing the permanent, induced, and eddy current fields that can be modeled relative to the inducing magnetic field of the earth with the Tolles-Lawson coefficients \cite{Gnadt2022a}:
\begin{align}
    \bfB_a &= \bfB_{\text{perm}} + \bfB_{\text{ind}} + \bfB_{\text{eddy}} \\
    &= \begin{pmatrix} a_1 \\ a_2 \\ a_3 \end{pmatrix} + \begin{pmatrix} \beta_1 & \beta_2 & \beta_3 \\ \beta_4 & \beta_5 & \beta_6 \\ \beta_7 & \beta_8 & \beta_9 \end{pmatrix} \bfB_e + \begin{pmatrix} c_1 & c_2 & c_3 \\ c_4 & c_5 & c_6 \\ c_7 & c_8 & c_9 \end{pmatrix} \dot{\bfB}_e 
\label{eq:tl_coef_def}
\end{align}
Since state-of-the-art vector magnetometers lack sufficient accuracy, a combination of vector and high-accuracy scalar magnetometers is typically carried onboard. The high-accuracy measurements of the scalar instrument have to be used for accurate calibration. A scalar magnetometer measures the absolute value of the total field,
\begin{align}
    B_t = |\bfB_e + \bfB_a|.
\end{align}
To extract the \ac{TL} coefficients from this nonlinear system, the approximation $B_e \ll B_a$ can be used. It is justified for aircraft fields much smaller than the Earth's magnetic field. This leads to
\begin{align}
    B_t \approx B_e + \bfB_a \cdot \frac{\bfB_e}{B_e},
\label{eq:Bt_approx}
\end{align}
where the term $\frac{\bfB_e}{B_e}$ is the direction of the Earth's magnetic field vector in the aircraft body frame. It is often expressed in terms of direction cosines as 
\begin{align}
    \hbfB_e \coloneqq \frac{\bfB_e}{B_e} = \begin{pmatrix}
        \cos \alpha \\ \cos \beta \\ \cos \gamma 
    \end{pmatrix} = \begin{pmatrix}
        \hat{B}_x \\ \hat{B}_y \\ \hat{B}_z
    \end{pmatrix} = \begin{pmatrix}
        B_{e,x}/B_e \\ B_{e,y}/B_e\\ B_{e,z}/B_e 
    \end{pmatrix}
\end{align}
There are several options for determining this term. 
Standard practice is to assume, again, that the aircraft field is small, $B_e \ll B_a$, replacing $\frac{\bfB_e}{B_e}$ by the measurement of a vector magnetometer
\begin{align}
    \hbfB_e \approx \hbfB_v = \frac{\bfB_v}{B_v} = \begin{pmatrix}
        \cos \alpha \\ \cos \beta \\ \cos \gamma 
    \end{pmatrix}.
\end{align}
This approximation introduces small errors that may become significant for large platform fields. While the vector magnetometer accuracy is not sufficient for calibration, it can determine the direction of $\bfB_e$ sufficiently well. As an example, an error of 500 nT in one component of the vector measurement of a \SI{50000}{\nano\tesla} field would only lead to an angle error of 0.01°.
Instead of relying on vector magnetometer measurements, it is generally also possible to determine $\hbfB_e$ using onboard inertial sensors. Assuming the magnetic background field direction is well represented by a core field model (WMM or IGRF-13) for calibration flights performed at high enough altitude, the aircraft attitude relative to the NED frame,  $\bfR_{nb}$, is determined from the inertial measurements and used to rotate $\hbfB_{\text{model}}$ into the aircraft body frame:
\begin{align}
     \hbfB^b_e = \bfR_{nb}^\top \hbfB^n_{\text{model}}(\bfp), 
\label{eq:att_cosines}
\end{align}
So far, all magnetic field vectors implicitly have been given in the aircraft body frame. We now introduce the superscripts $b$ and $n$ to identify if the magnetic field vector is expressed in body or NED coordinates. The IGRF magnetic vector is usually given in NED coordinates.
Inserting \cref{eq:tl_coef_def} into \cref{eq:Bt_approx} we obtain
\begin{align}
\begin{split}
    B_t &= B_e + (\bfB_{\text{perm}} + \bfB_{\text{ind}} + \bfB_{\text{eddy}}) \hbfB_e \\
    &= B_e + \begin{pmatrix} a_1 \\ a_2 \\ a_3 \end{pmatrix} \cdot \hbfB_e + \begin{pmatrix}
    \beta_1 & \beta_2 & \beta_3 \\ \beta_4 & \beta_5 & \beta_6 \\ \beta_7 & \beta_8 & \beta_9 \end{pmatrix}   \bfB_e \cdot \hbfB_e + \begin{pmatrix} c_1 & c_2 & c_3 \\ c_4 & c_5 & c_6 \\ c_7 & c_8 & c_9 \end{pmatrix} \dot{\bfB}_e \cdot \hbfB_e \\
    &= B_e + B^p_{\text{perm}} + B^p_{\text{ind}} + B^p_{\text{eddy}}
\label{eq:Bt_TL}
\end{split}
\end{align}
where $B^p_{\text{perm}}$, $B^p_{\text{ind}}$ and $B^p_{\text{eddy}}$ denote the projections of the permanent, induced and eddy current fields along the total magnetic field vector measured by the vector magnetometer, $\hbfB_e$ or $\hbfB_v$.
Let us now rewrite the term $B^p_{\text{ind}}$:
\begin{align}
\begin{split}
    B^p_{\text{ind}} &=  \begin{pmatrix}
    \beta_1 & \beta_2 & \beta_3 \\ \beta_4 & \beta_5 & \beta_6 \\ \beta_7 & \beta_8 & \beta_9 \end{pmatrix}  \bfB_e \cdot \hbfB_e \\
    &= \dfrac{1}{|\bfB_e|}\begin{pmatrix}
    \beta_1 B_x + \beta_2 B_y + \beta_3 B_z \\ \beta_4 B_x + \beta_5 B_y + \beta_6 B_z\\ \beta_7 B_x + \beta_8 B_y + \beta_9 B_z \end{pmatrix} \cdot \begin{pmatrix}
    B_x  \\ B_y \\  B_z \end{pmatrix} \\
     &= \dfrac{1}{|\bfB_e|} \left( \beta_1 B_x^2 + \beta_2 B_y B_x + \beta_3 B_z B_x + \beta_4 B_xB_y + \beta_5 B_y^2 + \beta_6 B_z B_y + \beta_7 B_x B_z + \beta_8 B_y B_z + \beta_9 B_z^2 \right)  \\
     &= \dfrac{1}{|\bfB_e|} \left( \beta_1 B_x^2 + (\beta_2 + \beta_4) B_x B_y + (\beta_3 + \beta_7) B_x B_z + \beta_5 B_y^2 + (\beta_6+\beta_8) B_y B_z + \beta_9 B_z^2\right) \\
     &=\begin{pmatrix}
    \beta_1 & (\beta_2 + \beta_4) & (\beta_3 + \beta_7) \\ 0 & \beta_5 & (\beta_6 + \beta_8) \\  0 & 0 & \beta_9 \end{pmatrix}  \bfB_e \cdot \hbfB_e. \\
\end{split}
\end{align}
By defining $b_1 = \beta_1 $, $b_2 = \beta_2 + \beta_4 $, $b_3 = \beta_3 + \beta_7 $, $b_4 = \beta_5$,  $b_5 = \beta_6 + \beta_8$, $b_6 = \beta_9$, we can write 
\begin{equation}
    B^p_{\text{ind}} =  \begin{pmatrix}
    b_1 & b_2 & b_3 \\ 0 & b_4 & b_5 \\ 0 & 0 & b_6 
    \end{pmatrix}  \bfB_e \cdot \hbfB_e 
\end{equation}

In practice, \cref{eq:Bt_TL} $\bfB_e$ and $\dot{\bfB}_e$ in \cref{eq:Bt_TL} are often replaced by $\bfB_t$ and $\dot{\bfB}_t$ such that absolute magnetometer measurements can be used to solve the equation:
\begin{align}
\begin{split}
    B_t &\approx B_e + \begin{pmatrix} a_1 \\ a_2 \\ a_3 \end{pmatrix} \hbfB_e + \begin{pmatrix} b_1 & b_2 & b_3 \\ 0 & b_4 & b_5 \\ 0 & 0 & b_6 \end{pmatrix} \bfB_t \hbfB_e + \begin{pmatrix} c_1 & c_2 & c_3 \\ c_4 & c_5 & c_6 \\ c_7 & c_8 & c_9 \end{pmatrix} \dot{\bfB}_t \hbfB_e \\
    &= B_e + \bfA_{\text{TL}} \bobeta_{\text{TL}}
\label{eq:Bt_TL_approx_2}
\end{split}
\end{align}
where 
\begin{align}
    \bobeta_{\text{TL}} = \begin{bmatrix}
        a_1 & a_2 & a_3 & b_1 & ... & b_5 & b_6 & c_1 & ... & c_9
    \end{bmatrix}^\top
\end{align}
and the Tolles-Lawson $\bfA_{\text{TL}}$-matrix is given by
\begin{align} \bfA_{\text{TL}} &=
    \begin{bmatrix}
        \hat{B}_x \\ \hat{B}_y \\ \hat{B}_z \\
        B_t \hat{B}_x^2 \\ B_t \hat{B}_x \hat{B}_y \\ B_t \hat{B}_x \hat{B}_z \\  B_t \hat{B}_y^2 \\  B_t \hat{B}_y \hat{B}_z \\ B_t \hat{B}_z^2 \\
        \hat{B}_x \dot{B}_x \\ \hat{B}_x \dot{B}_y \\ \hat{B}_x \dot{B}_z \\ \hat{B}_y \dot{B}_x \\ \hat{B}_y \dot{B}_y \\ \hat{B}_y \dot{B}_z \\ \hat{B}_z \dot{B}_x \\ \hat{B}_z \dot{B}_y \\ \hat{B}_z \dot{B}_z
    \end{bmatrix}^\top = B_t \begin{bmatrix}
        \hat{B}_x/B_t \\ \hat{B}_y/B_t \\ \hat{B}_z/B_t \\
        \hat{B}_x^2 \\\hat{B}_x \hat{B}_y \\ \hat{B}_x \hat{B}_z \\\hat{B}_y^2 \\\hat{B}_y \hat{B}_z \\ \hat{B}_z^2 \\
         \frac{\hat{B}_x \dot{B}_x}{B_t} \\ \frac{\hat{B}_x \dot{B}_y}{B_t} \\ \frac{\hat{B}_x \dot{B}_z}{B_t} \\ \frac{\hat{B}_y \dot{B}_x}{B_t} \\ \frac{\hat{B}_y \dot{B}_y}{B_t} \\ \frac{\hat{B}_y \dot{B}_z}{B_t} \\ \frac{\hat{B}_z \dot{B}_x}{B_t} \\ \frac{\hat{B}_z \dot{B}_y}{B_t} \\ \frac{\hat{B}_z \dot{B}_z}{B_t}
    \end{bmatrix}^\top =      \begin{bmatrix}
		\hat{B}_{x}            \\
		\hat{B}_{y}            \\
		\hat{B}_{z}            \\
		B_x \hat{B}_{x}        \\
		B_x \hat{B}_{y}        \\
		B_x \hat{B}_{z}        \\
		B_y \hat{B}_{y}        \\
		B_y \hat{B}_{z}        \\
		B_z \hat{B}_{z}        \\
		\dot{B}_x \hat{B}_{x} \\
		\dot{B}_x \hat{B}_{y} \\
		\dot{B}_x \hat{B}_{z} \\
		\dot{B}_y \hat{B}_{x} \\
		\dot{B}_y \hat{B}_{y} \\
		\dot{B}_y \hat{B}_{z} \\
		\dot{B}_z \hat{B}_{x} \\
		\dot{B}_z \hat{B}_{y} \\
		\dot{B}_z \hat{B}_{z}
	\end{bmatrix}^{\!\top}.
\label{eq:A_matrix}
\end{align}

We can introduce the index $i$ for a single measurement point, writing \cref{eq:Bt_TL_approx_2} as
\begin{align}
    B_{t,i} = B_{e,i} + \bfA_{\text{TL},i} \bobeta_{\text{TL}}.
\end{align}
For a series of $N$ measurements $i = 1,...N$ we get N equations
\begin{align}
    \bfB_{t,N} = \bfB_{e,N} + \bfA_{\text{TL},N\times18} \bobeta_{\text{TL}}
    \label{eq:TL_equation_system}
\end{align}
where $\bfB_{t,N}$ is a column vector containing the N scalar magnetometer measurements $B_t$ and $\bfA_{\text{TL},N\times18}$ is the $\bfA_{\text{TL}}$-matrix where each of the N rows contains the direction cosines $\hat{B}_x$,$\hat{B}_y$ and $\hat{B}_z$ and their time derivatives as well as the scalar magnetometer measurement for the $i$'th measurement. The time derivatives of the magnetic field components are calculated as finite differences:
\begin{equation}
\begin{split}
    \dot{B}_x &=  \left( B_{x,i} - B_{x,i-1} \right)/dt\\
    \dot{B}_y &= \left( B_{y,i} - B_{y,i-1} \right)/dt \\
    \dot{B}_z &= \left( B_{z,i} - B_{z,i-1} \right)/dt\
\end{split}
\end{equation}
This system of equations can now generally be solved for $\bobeta_{\text{TL}}$.
\subsection{Map-based calibration}
Knowing the external geomagnetic field $B_e$ makes the calibration process more robust. since it can be separated from the measurements, and the linear system in \cref{eq:TL_equation_system} can be solved directly. In solar quiet times, $B_e$ can generally be represented as a composition of the core field and local anomalies:
\begin{align}
    B_{e,i} = B_{\text{core}}(\bfp_i) + B_{\text{anomalies}}(\bfp_i)
\end{align}

In areas where the core field is assumed locally constant, $B_e$ may be assumed to be sufficiently well known from a core field model alone. In case of significant local variations, a map of the crust anomalies may improve calibration accuracy.

\subsection{Map-less calibration}
When $B_e$ is not known exactly, it is "removed" from the equation by applying a bandpass filter to the magnetic measurement data with a frequency range such that $\text{bpf}(B_e) \approx 0$. Then, the coefficients are obtained by solving: 
\begin{align}
    \text{bpf}(\bfB_{t,N}) = \text{bpf}(\bfA_{\text{TL}, N \times 18}) \bobeta_{\text{TL}}.
\end{align}
The range of the band-pass filter is chosen such as to optimally remove $B_e$ without loosing too much information on $B_a$. Typical frequency ranges can be found in \cite{jukic_applications_2024}, conservative bounds are, e.g., [0.002,1] Hz.
Determining the correct bandpass filtering is not a one-fits all solution and no guarantee for success. Depending on the magnetometer, too much or too little of the frequency content is removed. Overlapping frequency components of $B_a$ with the external field cannot be removed by this.
\subsection{Calibration using an accurate vector magnetometer}
Assuming that we have accurate measurements of $\bfB^b_t$ available, the approximation in eq. \cref{eq:Bt_approx} becomes unnecessary and we obtain, instead, three equations for each measurement point $i$:
\begin{align}
    \bfB^b_{t,i} = \bfB^b_{e,i} + \begin{pmatrix} a_1 \\ a_2 \\ a_3 \end{pmatrix} + \begin{pmatrix}
    \beta_1 & \beta_2 & \beta_3 \\ \beta_4 & \beta_5 & \beta_6 \\ \beta_7 & \beta_8 & \beta_9 \end{pmatrix}  \bfB^b_{e,i} +    \begin{pmatrix} c_1 & c_2 & c_3 \\ c_4 & c_5 & c_6 \\ c_7 & c_8 & c_9 \end{pmatrix}\dot{\bfB}^b_{e,i}
\end{align}

Now $\bfB_{t,i}$ is the vector magnetometer measurement. We have, again, two alternatives to determine $\bfB_{e,i}$. In map-based calibration, we add the projection of the magnetic anomaly field along the core field to the core field:
\begin{align}
    \bfB^b_{e,i} \approx \bfR_{nb,i}^\top ( \bfB^n_{\text{model}}(\bfp_i) + B_{\text{map}}(\bfp_i) \hbfB^n_{\text{model}}(\bfp_i)).
\end{align}
For map-less calibration, when the calibration flight is at high enough altitude, one can just use
\begin{align}
    \bfB^b_{e,i} \approx \bfR_{nb,i}^\top \bfB^n_{\text{model}}(\bfp_i).
\end{align}
The final system of $3N$ equations to be solved for the 18 coefficients via regression is then

\begin{equation} 
\bfB^b_{t,3N} = \bfB^b_{e,3N} + \left(N \bfI \otimes \begin{pmatrix} a_1 \\ a_2 \\ a_3 \end{pmatrix}\right) + \left(N \bfI \otimes \begin{pmatrix}
    \beta_1 & \beta_2 & \beta_3 \\ \beta_4 & \beta_5 & \beta_6 \\ \beta_7 & \beta_8 & \beta_9 \end{pmatrix}\right) \bfB^b_{e,3N} + \left(N \bfI \otimes \begin{pmatrix} c_1 & c_2 & c_3 \\ c_4 & c_5 & c_6 \\ c_7 & c_8 & c_9 \end{pmatrix}\right) \dot{\bfB}^b_{e,3N} 
\end{equation}

\subsection{Jacobian of the Tolles-Lawson Model}
\label{ssec:jac_TL}
When one wants to use the Tolles-Lawson Model for Online-Calibration, it is included as a term in the Kalman Filter measurement equation for the magnetic field. Its parameters are updated as part of the state vector simultaneously with navigation. In this case, the partial derivatives of the model function with respect to its vector inputs and parameters need to be calculated for the measurement Jacobian.

The derivative with respect to the Tolles-Lawson parameters is simply given by
\begin{align}
    \frac{\partial (\bfA_{\text{TL}} \bobeta_{\text{TL}})}{\partial \bobeta_{\text{TL}}} = \bfA_{\text{TL}}.
\end{align}
The derivatives with respect to the vector magnetometer measurements (using the scalar magnetometer measurement as input for $B_t$) are

\begin{subequations}
    \begin{align}
\frac{\partial (\bfA_\text{TL} \bobeta_\text{TL})}{\partial B_x} &= \frac{a_1}{B_t} + \left( 2 \hat{B}_x b_1 + \hat{B}_y b_2 + \hat{B}_z b_3 \right) \nonumber \\
&\quad + \frac{1}{B_t} \left( \dot{B}_x c_1 + \dot{B}_y c_4 + \dot{B}_z c_7 \right) \\[1em]
\frac{\partial (\bfA_\text{TL} \bobeta_\text{TL})}{\partial B_y} &= \frac{a_2}{B_t} + \left( \hat{B}_x b_2 + 2 \hat{B}_y b_4 + \hat{B}_z b_5 \right) \nonumber \\
&\quad + \frac{1}{B_t} \left( \dot{B}_x c_2 + \dot{B}_y c_5 + \dot{B}_z c_8 \right) \\[1em]
\frac{\partial (\bfA_\text{TL} \bobeta_\text{TL})}{\partial B_z} &= \frac{a_3}{B_t} + \left( \hat{B}_x b_3 + \hat{B}_y b_5 + 2 \hat{B}_z b_6 \right) \nonumber \\
&\quad + \frac{1}{B_t} \left( \dot{B}_x c_3 + \dot{B}_y c_6 + \dot{B}_z c_9 \right).
\end{align}
\label{eq:del_TL_del_vec}
\end{subequations}

%% file: appendix/equiv_ekf_ng.tex
\section{Mathematical Equivalence of EKF and Online Natural Gradient}
\label{ssec:equivalence_ekf_ng}

This section provides a brief mathematical derivation of the equivalence of the \ac{EKF} update for neural network training to an online, stochastic \ac{NG} descent as shown in \cite{Ollivier2018}. We show that the \ac{EKF} update \eqref{eq:kf_update_state_suppl} is an implementation of an online Newton method on the log-likelihood, where the preconditioner is the inverse of the total accumulated Fisher Information Matrix (\ac{FIM}).

The EKF standard prediction-correction equations for a discretized static system with system state $\bfx$ are:

\begin{enumerate}
    \item Prediction (propagation in time):
    \begin{align}
    \bfx_{t|t-1} &\leftarrow \bfx_{t-1} \\
    \covP_{t|t-1} &\leftarrow \covP_{t-1} + \covQ_d
    \end{align}
    \item Correction (measurement update):
    \begin{subequations}
    \begin{align}
    \bfK_t &= \covP_{t|t-1} \bfH^\top (\bfH \covP_{t|t-1} \bfH^\top + \covR)^{-1} \\
    \bfx_{t} &= \bfx_{t|t-1} + \bfK_t (\bfz_t - \bfh(\bfx_{t|t-1})) \label{eq:kf_update_state_suppl} \\
    \covP_t &= (\bfI - \bfK_t \bfH) \covP_{t|t-1} \label{eq:ekf_update_covariance_suppl},
    \end{align}
    \label{eq:ekf_update_suppl}
    \end{subequations}
\end{enumerate}

\subsection{Basics}
\paragraph*{Negative Log-Likelihood (NLL) as the Loss}
The \ac{EKF} update \eqref{eq:ekf_update_suppl} is derived from a Bayesian, probabilistic perspective. The loss function it minimizes is the negative log-likelihood (NLL) of the observation $\bfz_t$ given the parameters $\bfx_{\text{NN}}$, which are assumed to be corrupted by Gaussian noise with covariance $\covR$ (as in \eqref{eq:ekf_update_suppl}).

The loss for a single new observation $\bfz_t$ is:
\begin{align}
    L_t(\bfx_{\text{NN}}) &= -\log p(\bfz_t | \bfx_{\text{NN}}) \\
    &= C + \frac{1}{2} \left( \bfz_t - \bfh(\bfx_{\text{NN}}) \right)^\top \covR^{-1} \left( \bfz_t - \bfh(\bfx_{\text{NN}}) \right)
\end{align}
where $C$ is a constant.

\paragraph*{Ordinary Gradient of the NLL}
The ordinary gradient of this loss function with respect to the parameters $\bfx_{\text{NN}}$ is (using the chain rule):
\begin{align}
    \nabla_{\bfx_{\text{NN}}} L_t 
    &= \frac{\partial L_t}{\partial \bfx_{\text{NN}}} = \left( \frac{\partial \bfh(\bfx_{\text{NN}})}{\partial \bfx_{\text{NN}}} \right)^\top \frac{\partial L_t}{\partial \bfh} \\
    &= - \left( \frac{\partial \bfh(\bfx_{\text{NN}})}{\partial \bfx_{\text{NN}}} \right)^\top \covR^{-1} (\bfz_t - \bfh(\bfx_{\text{NN}})) \\
    &= - \bfH_t^\top \covR^{-1} (\bfz_t - \bfh(\bfx_{\text{NN}}))
    \label{eq:loss_gradient}
\end{align}
where $\bfH_t = \frac{\partial \bfh}{\partial \bfx_{\text{NN}}}$ is the measurement Jacobian.

\paragraph*{Fisher Information Matrix (FIM)}
The \ac{FIM} $\bfJ_t$ for this observation is the expected squared gradient. For a Gaussian model, the \ac{FIM} (and the Hessian of the NLL, using the Gauss-Newton approximation) is:
\begin{align}
    \bfJ_t = \mathbb{E} \left[ (\nabla_{\bfx_{\text{NN}}} L_t) (\nabla_{\bfx_{\text{NN}}} L_t)^\top \right] \approx \bfH_t^\top \covR^{-1} \bfH_t
\end{align}

\subsection{EKF Update as Gradient Descent}

The EKF update can be rewritten from its Kalman Gain form to its Gradient Descent form. The key identity (derived in Lemma 8 \cite{Ollivier2018}) is:
\begin{equation}
    \bfK_t \covR = \covP_t \bfH_t^\top
\end{equation}
Assuming $\covR$ is invertible, $\bfK_t = \covP_t \bfH_t^\top \covR^{-1}$. We substitute this into the standard state update:
\begin{align}
    \bfx_t &= \bfx_{t|t-1} + \bfK_t (\bfz_t - \bfh(\bfx_{t|t-1})) \\
    &= \bfx_{t|t-1} + \left( \covP_t \bfH_t^\top \covR^{-1} \right) (\bfz_t - \bfh(\bfx_{t|t-1})) \\
    &= \bfx_{t|t-1} + \covP_t \left( \bfH_t^\top \covR^{-1} (\bfz_t - \bfh(\bfx_{t|t-1})) \right) \label{eq:ekf_gradient_form}
\end{align}
Comparing this to the ordinary gradient $\nabla L_t$ from \cref{eq:loss_gradient}, we find:
\begin{equation}
    \bfx_t = \bfx_{t|t-1} - \covP_t (\nabla_{\bfx_{\text{NN}}} L_t)
    \label{eq:ekf_update_loss}
\end{equation}
This shows that the EKF update is an online gradient descent preconditioned by the covariance $\covP_t$.

\subsection{EKF as Online Natural Gradient and the connection to the Learning Rates}
\label{sssec:ekf_ng_learningrates}
This section shows that the EKF is always equivalent to an \ac{NG} descent. The value of $\covQ$ changes the resulting learning rate.

\paragraph*{The EKF/NG Equivalence}

The \ac{NG} update is:
\begin{equation}
    \bfx_t = \bfx_{t-1} - \eta_t (\bfJ_t^{-1}) (\nabla_{\bfx_{\text{NN}}} L_t)
\end{equation}
Comparing this to \cref{eq:ekf_update_loss}, the \ac{EKF} is identical to an \ac{NG} descent where the covariance $\covP_t$ is the (scaled) preconditioner $\eta_t \bfJ_t^{-1}$. The learning rate is the evolving covariance matrix $\covP_t$. The tuning parameters ($\covP_0$, $\covQ$, $\covR$) control how $\covP_t$ evolves.

\paragraph*{The Initial Learning Rate (Transient Phase)}
$\covP_0$ acts as the Bayesian prior, or the initial learning rate. For the very first update step ($t=1$), the information matrix is $\covP_1^{-1} = \covP_0^{-1} + \bfJ_1$.
\begin{itemize}
    \item \textbf{Large $\covP_0$ (Weak Prior):} $\covP_0^{-1} \to \bfzero$, so $\covP_1 \approx \bfJ_1^{-1}$. The filter learns aggressively (high initial $\eta$).
    \item \textbf{Small $\covP_0$ (Strong Prior):} $\covP_0^{-1} \to \infty$, so $\covP_1 \approx \covP_0$. The filter learns slowly, almost ``ignoring'' the data (low initial $\eta$).
\end{itemize}

\paragraph*{The Long-Term Learning Rate (Steady-State Phase)}
After the initial transient phase, the behavior of $\covP_t$ is governed by $\covQ$ and $\covR$.

\subparagraph{Case A: The Decaying Rate ($\covQ = 0$)}
As a side note, if $\covQ=0$, the EKF optimally averages all past data. This is proven in \cite[~Th.2]{Ollivier2018} to be equivalent to an \ac{NG} with a decaying learning rate $\eta_t = 1/(t+1)$.

\subparagraph{Case B: The Constant Rate ($\covQ > 0$)}
In our application, we are interested in a non-decaying learning rate, which is achieved by setting $\covQ > \bfzero$. The $\covQ/\covR$ ratio controls this long-term or steady-state learning rate ($\covP^*$). This is found by solving the Discrete Algebraic Riccati Equation (DARE) at steady-state ($\covP_t = \covP_{t-1} = \covP^*$), where the uncertainty added by $\covQ_d$ is balanced by the information from $\covR$:
\begin{equation}
\covP^{*-1} = (\covP^* + \covQ_d)^{-1} + \bfH_t^\top \covR^{-1} \bfH_t
\end{equation}

Analyzing this balance shows:
\begin{itemize}
    \item If $\covQ_d \uparrow$ (numerator), $\covP^* \uparrow$, which means $\eta^* \uparrow$ (Higher Learning Rate).
    \item If $\covR \uparrow$ (denominator), $\covP^* \downarrow$, which means $\eta^* \downarrow$ (Lower Learning Rate).
\end{itemize}
This confirms that tuning the $\covQ/\covR$ ratio is the principled mechanism for setting the stable, long-term learning rate $\eta^*$ for our $\covQ > \bfzero$ case. The common practice is to fix $\covR$ and tune $\covQ$ to achieve the desired $\covP^*$ (learning rate).

\subsection{Summary}
The EKF is equivalent to an online Natural Gradient descent, in particular: 
\begin{itemize}
    \item The EKF update is a preconditioned gradient descent.
    \item The covariance matrix $\covP_t$ is the (scaled) inverse of the total Fisher Information Matrix.
    \item The initial covariance $\covP_0$ is the initial learning rate.
    \item The ratio of $\covQ/\covR$ controls the steady-state learning rate.
\end{itemize}

%% file: appendix/nn_jacobian.tex
\section{The Neural Network Jacobian}
\label{sec:nn_jacobian}
The Jacobian of the network with respect to its parameters, $\frac{\partial \tilde{g}}{\partial \boLambda_{\text{NN}}}$, is calculated by recursively applying the chain rule. We assume the parameters (weight matrices $\bfW$ and bias vectors $\bfb$) of a network with $L$ layers are flattened and concatenated into the state vector as follows:
 \begin{align}
    \boLambda_{\text{NN}} = \left[ 
    \text{vec}(\mathbf{W}^{(1)}), 
    \mathbf{b}^{(1)}, 
    \dots, 
    \text{vec}(\mathbf{W}^{(L)}), 
    \mathbf{b}^{(L)} 
 \right]^\top.
 \end{align}

To illustrate how the network derivatives are formed, we can expand the partial derivatives for a simple two-layer network using the chain rule. Assuming a hidden layer with $N_h$ neurons, a feature input $\bophi_{\text{i}} \in \mathbb{R}^{N_f}$, and a $\tanh(\cdot)$ activation function, the network output is:
\begin{align}
    \tilde{g}(\bophi_{\text{i}}, \boLambda_{\text{NN}}) = \sum_{i=1}^{N_h} \left( w_{2,i} \cdot \tanh \left( \sum_{k=1}^{N_f} w_{1,i,k} \tilde{\phi}_k + b_{1,i} \right) \right) + b_2
\end{align}
Letting $\zeta_i$ be the pre-activation of the i-th hidden neuron, the partial derivatives of the network output $\tilde{g}$ with respect to its parameters $\boLambda$ are:
\begin{align}
\frac{\partial \tilde{g}}{\partial w_{1,i,k}} &= \frac{\partial \tilde{g}}{\partial \tanh(\zeta_i)} \frac{\partial \tanh(\zeta_i)}{\partial \zeta_i} \frac{\partial \zeta_i}{\partial w_{1,i,k}} = w_{2,i} \cdot (1 - \tanh^2(\zeta_i)) \cdot \tilde{\phi}_k, \\[1em]
\frac{\partial \tilde{g}}{\partial b_{1,i}} &= \frac{\partial \tilde{g}}{\partial \tanh(\zeta_i)} \frac{\partial \tanh(\zeta_i)}{\partial \zeta_i} \frac{\partial \zeta_i}{\partial b_{1,i}} = w_{2,i} \cdot (1 - \tanh^2(\zeta_i)), \\[1em]
\frac{\partial \tilde{g}}{\partial w_{2,i}} &= \tanh(\zeta_i), \\[1em]
\frac{\partial \tilde{g}}{\partial b_{2}} &= 1
\label{eq:derivatives}
\end{align}

While illustrative, manually deriving these derivatives for deep or complex networks is impractical and can be automated in the implementation. 

%% file: appendix/crlb_odom.tex
\section{Cramér-Rao Lower Bound Analysis of the Odometry-Model}
\label{sec:CRLB_suppl}
To establish a theoretical benchmark, we calculated the \ac{CRLB} as the inverse of the \ac{FIM}, $\bfJ$. The \ac{FIM} is computed recursively via
\begin{subequations}
\begin{align}
    \bfJ_{t|t-1} &= (\bfF_{t-1} \bfJ_{t-1}^{-1} \bfF_{t-1}^T + \covQ_{t-1})^{-1} \label{eq:fim_pred} \\
    \bfJ_t &= \bfJ_{t|t-1} + \bfH_t^T \covR_t^{-1} \bfH_t. \label{eq:fim_update}
\end{align}
\end{subequations}
Since true neural network parameters do not exist, we evaluated $\bfH_t$ using the filter's current parameter estimates, yielding an estimator-informed CRLB that quantifies the filter's theoretical self-consistency.

\begin{figure}[tbh!]
    \centering
    \begin{subfigure}[b]{0.45\textwidth}
        \centering
        \includegraphics[width=\textwidth]{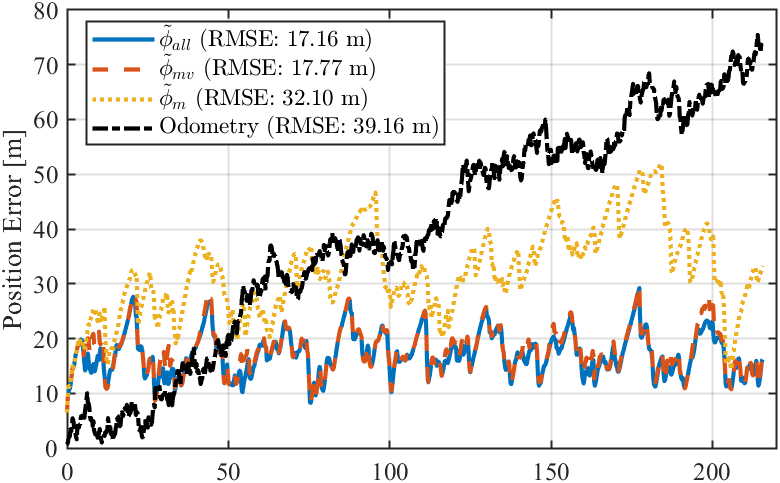}
        \caption{}
        \label{fig:crlb_a}
    \end{subfigure}
    \hfill 
    \begin{subfigure}[b]{0.45\textwidth}
        \centering
        \includegraphics[width=\textwidth]{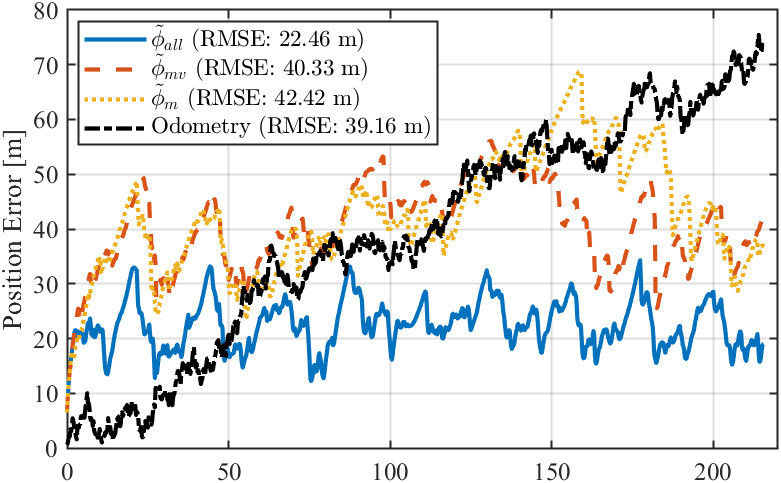}
        \caption{}
        \label{fig:crlb_b}
    \end{subfigure}
    
    \caption{CRLBs along the lawnmower-trajectory for (a) small $N_h = 4$ and (b) large $N_h = 32$.}
    \label{fig:toymodel_crlb}
\end{figure}

The visualization of these bounds in \cref{fig:toymodel_crlb} reveals that the compact architecture ($N_h=4$, \cref{fig:crlb_a}) maintains a low theoretical error bound, and the larger network ($N_h=32$, \cref{fig:crlb_b}) exhibits a significantly higher uncertainty floor, particularly for the configurations with fewer features, ($\tilde{\phi}_\text{m}$ and $\tilde{\phi}_\text{mv}$). This demonstrates that increasing model complexity without additional information dilutes the observability of the system. This is the information-theoretic analog to overfitting. A significant portion of the information gained from each measurement is consumed by the estimation of the high-dimensional parameter space, fundamentally raising the minimum achievable variance for the navigation states themselves.

%% file: appendix/jacobian_nn-tl.tex
\section{Jacobian of the combined NN-TL calibration model with Respect to the Vector Magnetometer Input}
\label{ssec:jacobian_mvec}

The components of $\frac{\partial \bfh}{\partial \bfm}$ are given, analogous to \cite{Canciani2022} and \cref{ssec:jac_TL}, as
\begin{subequations}
\begin{align}
    \frac{\partial \bfh}{\partial m_x} = \frac{\bfx_{TL_1}}{m} + \left( 2 \frac{m_x}{m} \bfx_{TL_4} + \frac{m_y}{m} \bfx_{TL_5} + \frac{m_z}{m} \bfx_{TL_6} + \frac{\dot{m}_x}{m} \bfx_{TL_{10}} + \frac{\dot{m}_y}{m} \bfx_{TL_{11}} + \frac{\dot{m}_z}{m} \bfx_{TL_{12}} \right) + \frac{\partial \tilde{g}_{\text{res}}}{\partial m_x} \\
    \frac{\partial \bfh}{\partial m_y} = \frac{\bfx_{TL_2}}{m} + \left( \frac{m_x}{m} \bfx_{TL_5} + 2 \frac{m_y}{m} \bfx_{TL_7} + \frac{m_z}{m} \bfx_{TL_8} + \frac{\dot{m}_x}{m} \bfx_{TL_{13}} + \frac{\dot{m}_y}{m} \bfx_{TL_{14}} + \frac{\dot{m}_z}{m} \bfx_{TL_{15}} \right) + \frac{\partial \tilde{g}_{\text{res}}}{\partial m_y} \\
    \frac{\partial \bfh}{\partial m_z} = \frac{\bfx_{TL_3}}{m} + \left( \frac{m_x}{m} \bfx_{TL_6} + \frac{m_y}{m} \bfx_{TL_8} + 2 \frac{m_z}{m} \bfx_{TL_9} + \frac{\dot{m}_x}{m} \bfx_{TL_{16}} + \frac{\dot{m}_y}{m} \bfx_{TL_{17}} + \frac{\dot{m}_z}{m} \bfx_{TL_{18}} \right) + \frac{\partial \tilde{g}_{\text{res}}}{\partial m_z},
\end{align}
\label{eq:dh_dmvec}
\end{subequations}
where $m_i$ are the components of the vector magnetometer measurement, and $m$ is the scalar magnetometer measurement. The terms $\frac{\partial \tilde{g}_{\text{res}}}{\partial \bfm}$ are calculated by differentiating the \ac{NN} with respect to the vector magnetometer input.

%% file: appendix/additional_results.tex
\section{Additional results}
\label{ssec:suppl_results}
%
%
%
For completeness we report the detailed results for the remaining magnetometers 1, 3, 4 and 5 contained in the MagNav Dataset \cite{gnadt_daf-mit_2023}, as well as additional detail plots for Magnetometer 2 along the flight line 1007.06 presented in the main paper. We further show results along a second flight line 1003 using Magnetometer 3.
\subsection{Magnetometer 1}
\label{ssec:mag1_suppl}
\begin{figure}[tbh!]
    \centering
    \begin{subfigure}[t]{0.48\linewidth}
        \centering
        \includegraphics[width=\linewidth]{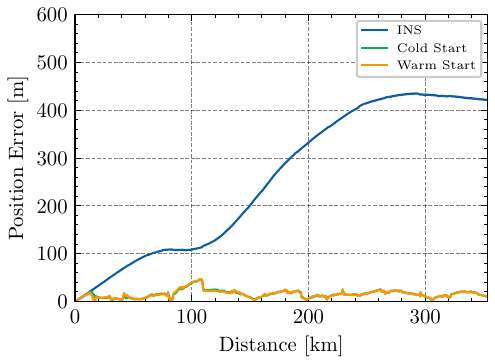}
        \caption{}
        \label{fig:cold_warm_nn_tl_mag1}
    \end{subfigure}
    \hfill
    \begin{subfigure}[t]{0.48\linewidth}
        \centering
        \includegraphics[width=\linewidth]{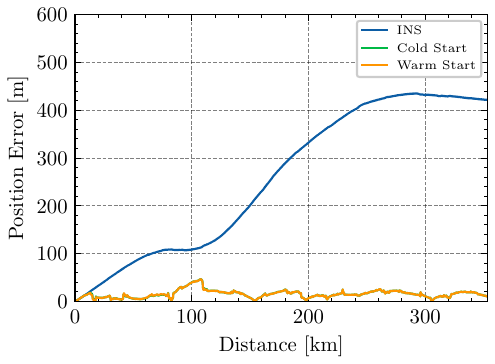}
        \caption{}
        \label{fig:cold_warm_tl_mag1}
    \end{subfigure}

    \caption{Evolution of the positioning accuracy in cold and warm start scenarios using Magnetometer 1 ($N_h = 5$). (a) Our hybrid approach combining \ac{TL} and \ac{NN} calibration (\ac{DRMS} cold-start \SI{17}{\meter} / warm-start \SI{17}{\meter}) (b) \ac{TL} online calibration (\ac{DRMS} cold-start \SI{17}{\meter} / warm-start \SI{17}{\meter}).}
    \label{fig:cold_warm_mag1_comparison}
\end{figure}

\begin{figure}[tbh!]
    \centering
    \begin{subfigure}[t]{0.48\linewidth}
        \centering
        \includegraphics[width=\linewidth]{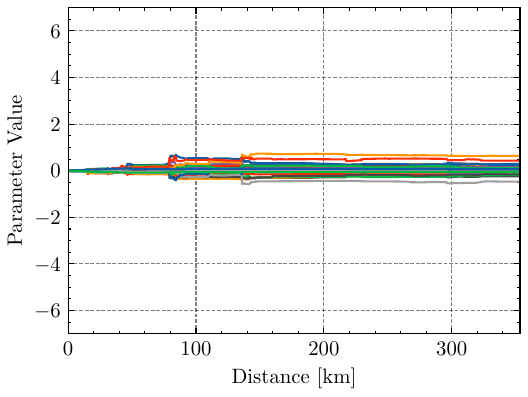}
        \caption{}
        \label{fig:nn_cold_start_mag1}
    \end{subfigure}
    \hfill
    \begin{subfigure}[t]{0.48\linewidth}
        \centering
        \includegraphics[width=\linewidth]{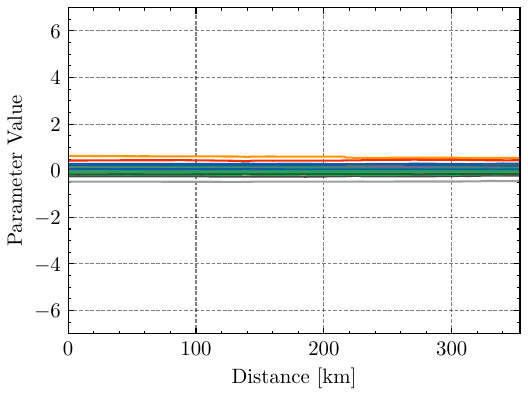}
        \caption{}
        \label{fig:nn_warm_start_mag1}
    \end{subfigure}
    \hfill
    \begin{subfigure}[t]{0.48\linewidth}
        \centering
        \includegraphics[width=\linewidth]{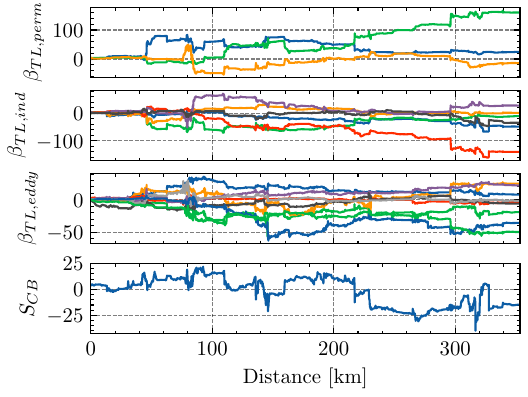}
        \caption{}
        \label{fig:tl_parameters_cold_mag1}
    \end{subfigure}
    \hfill
    \begin{subfigure}[t]{0.48\linewidth}
        \centering
        \includegraphics[width=\linewidth]{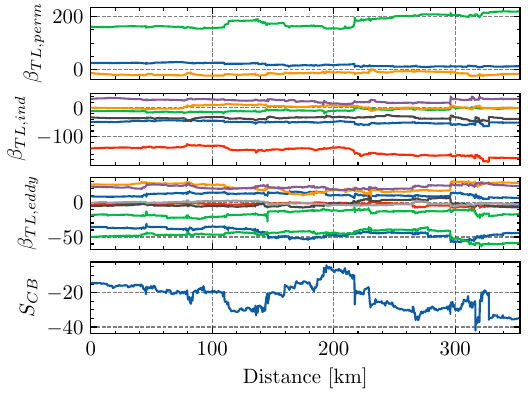}
        \caption{}
        \label{fig:tl_parameters_warm_mag1}
    \end{subfigure}

    \caption{Evolution of \ac{NN} (top) and \ac{TL} model (bottom) parameters using Magnetometer 1 ($N_h = 5$), comparing (a), (c) cold start and (b), (d) warm start.}
    \label{fig:tl_parameters_mag1}
\end{figure}

\begin{figure}[tbh!]
    \centering
    \begin{subfigure}[t]{0.48\linewidth}
        \centering
        \includegraphics[width=\linewidth]{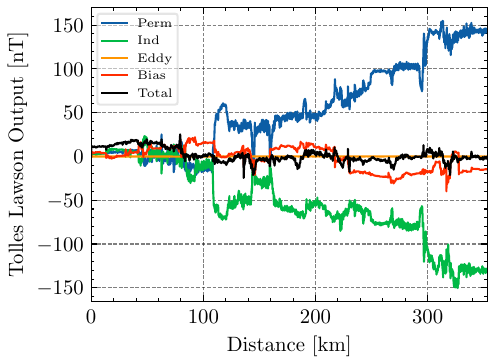}
        \caption{}
        \label{fig:tl_components_mag1}
    \end{subfigure}
    \hfill
    \begin{subfigure}[t]{0.48\linewidth}
        \centering
        \includegraphics[width=\linewidth]{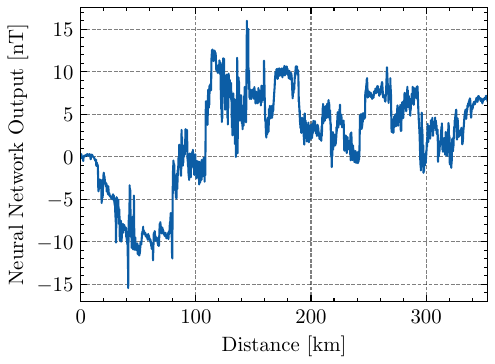}
        \caption{}
        \label{fig:nn_output_mag1}
    \end{subfigure}

    \caption{Comparison between the (a) \ac{TL} model compensation components (including the bias state $S_{CB}$) and the (b) learned \ac{NN} output using Magnetometer 1 in the cold start scenario.}
    \label{fig:tl_vs_nn_output_mag1}
\end{figure}

\begin{figure}[tbh!]
    \centering
    \begin{subfigure}[t]{0.48\linewidth}
        \centering
        \includegraphics[width=\linewidth]{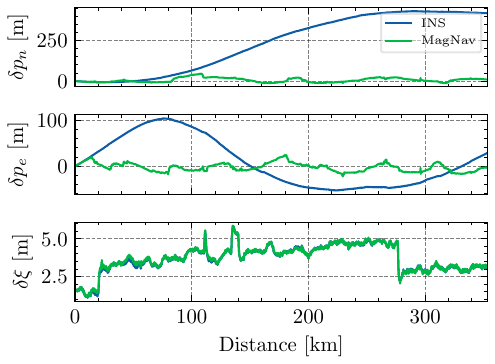}
        \caption{}
        \label{fig:pos_errors_cold_mag1}
    \end{subfigure}
    \hfill
    \begin{subfigure}[t]{0.48\linewidth}
        \centering
        \includegraphics[width=\linewidth]{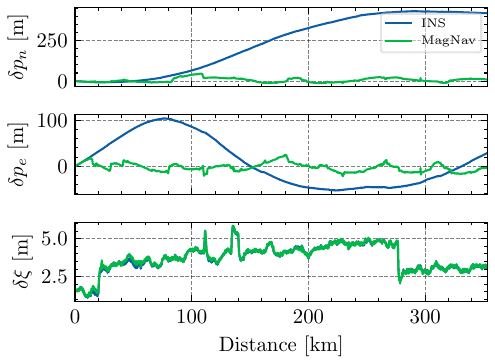}
        \caption{}
        \label{fig:pos_errors_warm_mag1}
    \end{subfigure}

    \caption{Comparison of INS and MagNav position errors (north, east, and down) for Magnetometer 1 in (a) our hybrid \ac{TL}+\ac{NN} cold-start and (b) the \ac{TL}-only cold-start scenarios.}
    \label{fig:pos_errors_cold_warm_mag1}
\end{figure}

Here we show the results of our MagNav filter for flight line 1007.06 using the almost noise-free reference Magnetometer 1. \cref{fig:cold_warm_mag1_comparison} shows the evolution of the position errors in the cold- and warm start scenarios. \cref{fig:tl_parameters_mag1} shows the estimated parameters of the Tolles-Lawson and \ac{NN} model over time. \cref{fig:tl_vs_nn_output_mag1} shows their output for the correction of the magnetic field. \cref{fig:pos_errors_cold_warm_mag1} is a detailed plot of the component-wise position errors.

\clearpage
\subsection{Magnetometer 2}
\label{ssec:mag2_suppl}
\begin{figure}[tbh!]
    \centering
    \begin{subfigure}[t]{0.48\linewidth}
        \centering
        \includegraphics[width=\linewidth]{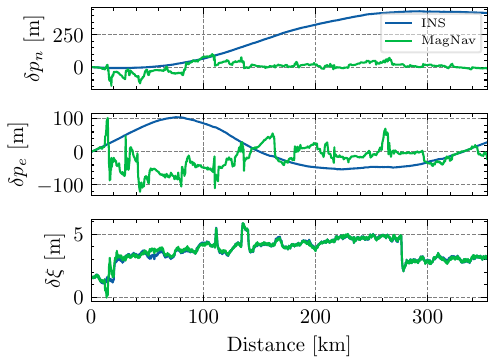}
        \caption{}
        \label{fig:pos_errors_cold_mag2}
    \end{subfigure}
    \hfill
    \begin{subfigure}[t]{0.48\linewidth}
        \centering
        \includegraphics[width=\linewidth]{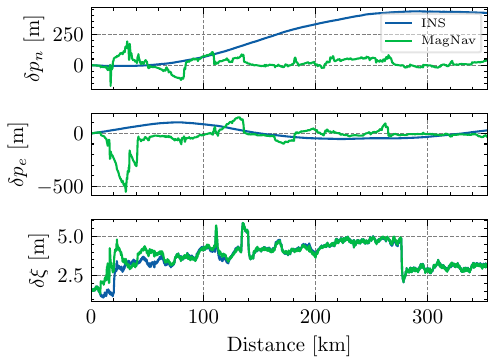}
        \caption{}
        \label{fig:pos_errors_warm_mag2}
    \end{subfigure}

    \caption{Comparison of INS and MagNav position errors (north, east, and down) for Magnetometer 2 in (a) our hybrid \ac{TL}+\ac{NN} cold-start and (b) the \ac{TL}-only cold-start scenarios.}
    \label{fig:pos_errors_cold_warm_mag2}
\end{figure}
Figure \ref{fig:pos_errors_cold_warm_mag2} compares the north, east, and down position errors for the TL+NN and TL-only cold-start runs using magnetometer 2. Across all three axes, the TL+NN configuration exhibits smaller transient errors and faster convergence than the TL-only case. In the north channel, the TL+NN filter stabilizes faster, whereas the TL-only filter shows larger oscillations and a slower decay of the initial divergence. The east error shows the most pronounced improvement: TL-only exhibits an early large error spike exceeding \SI{500}{\meter}, while TL+NN remains tightly bounded throughout the flight. The down error is well controlled in both cases, but TL+NN yields slightly reduced oscillations and fewer sharp peaks. Overall, the TL+NN filter provides improved transient suppression and smoother error evolution compared to the TL-only configuration.
\clearpage
\subsection{Magnetometer 3}
\label{ssec:mag3_suppl}

\begin{figure}[tbh!]
    \centering
    \begin{subfigure}[t]{0.48\linewidth}
        \centering
        \includegraphics[width=\linewidth]{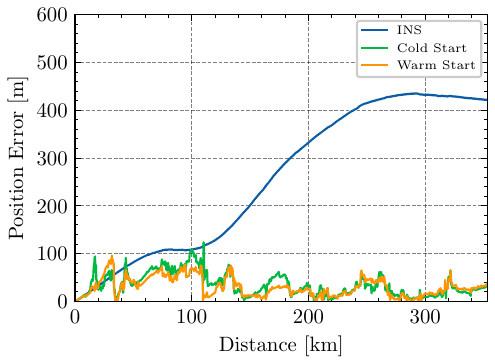}
        \caption{}
        \label{fig:cold_warm_nn_tl_mag3}
    \end{subfigure}
    \hfill
    \begin{subfigure}[t]{0.48\linewidth}
        \centering
        \includegraphics[width=\linewidth]{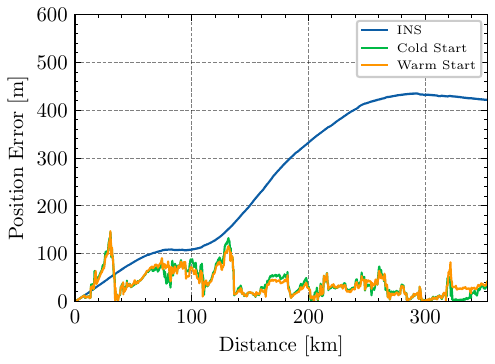}
        \caption{}
        \label{fig:cold_warm_tl_mag3}
    \end{subfigure}

    \caption{Evolution of the positioning accuracy in cold and warm start scenarios using Magnetometer 3 ($N_h = 2$). (a) Hybrid approach combining \ac{TL} and \ac{NN} calibration (\ac{DRMS} cold-start \SI{42}{\meter} / warm-start \SI{36}{\meter}) (b) \ac{TL} online calibration (\ac{DRMS} cold-start \SI{46}{\meter} / warm-start \SI{45}{\meter}).}
    \label{fig:cold_warm_mag3_comparison}
\end{figure}

\begin{figure}[tbh!]
    \centering
    \begin{subfigure}[t]{0.48\linewidth}
        \centering
        \includegraphics[width=\linewidth]{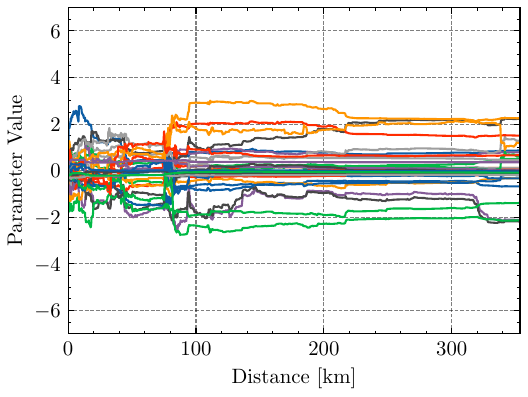}
        \caption{}
        \label{fig:nn_cold_start_mag3}
    \end{subfigure}
    \hfill
    \begin{subfigure}[t]{0.48\linewidth}
        \centering
        \includegraphics[width=\linewidth]{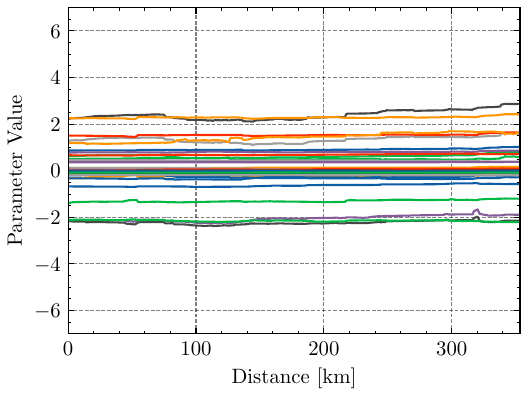}
        \caption{}
        \label{fig:nn_warm_start_mag3}
    \end{subfigure}
    \hfill
    \begin{subfigure}[t]{0.48\linewidth}
        \centering
        \includegraphics[width=\linewidth]{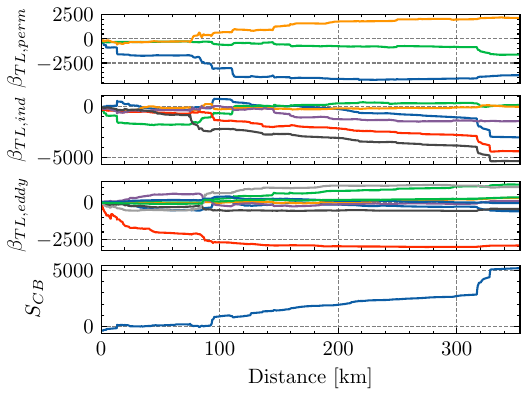}
        \caption{}
        \label{fig:tl_parameters_cold_mag3}
    \end{subfigure}
    \hfill
    \begin{subfigure}[t]{0.48\linewidth}
        \centering
        \includegraphics[width=\linewidth]{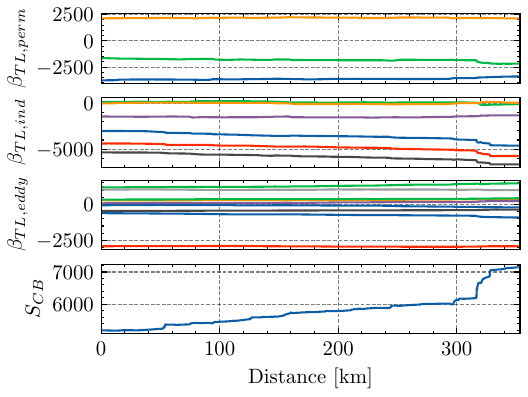}
        \caption{}
        \label{fig:tl_parameters_warm_mag3}
    \end{subfigure}

    \caption{Evolution of \ac{NN} (top) and \ac{TL} model (bottom) parameters using Magnetometer 3 ($N_h = 5$), comparing (a), (c) cold start and (b), (d) warm start.}
    \label{fig:tl_parameters_mag3}
\end{figure}

\begin{figure}[tbh!]
    \centering
    \begin{subfigure}[t]{0.48\linewidth}
        \centering
        \includegraphics[width=\linewidth]{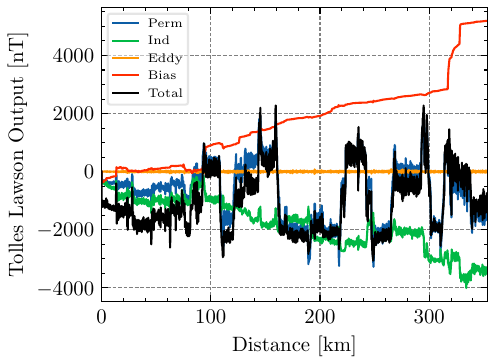}
        \caption{}
        \label{fig:tl_components_mag3}
    \end{subfigure}
    \hfill
    \begin{subfigure}[t]{0.48\linewidth}
        \centering
        \includegraphics[width=\linewidth]{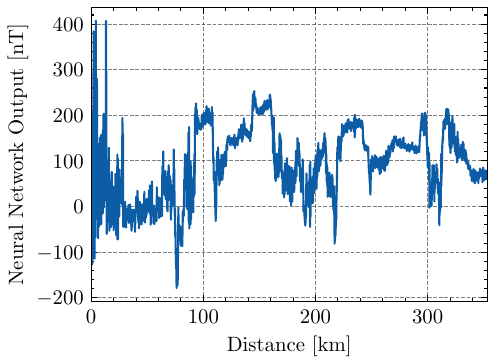}
        \caption{}
        \label{fig:nn_output_mag3}
    \end{subfigure}

    \caption{Comparison between the (a) \ac{TL} model compensation components (including the bias state $S_{CB}$) and the (b) learned \ac{NN} output using Magnetometer 3 in the cold start scenario.}
    \label{fig:tl_vs_nn_output_mag3}
\end{figure}

\begin{figure}[tbh!]
    \centering
    \begin{subfigure}[t]{0.48\linewidth}
        \centering
        \includegraphics[width=\linewidth]{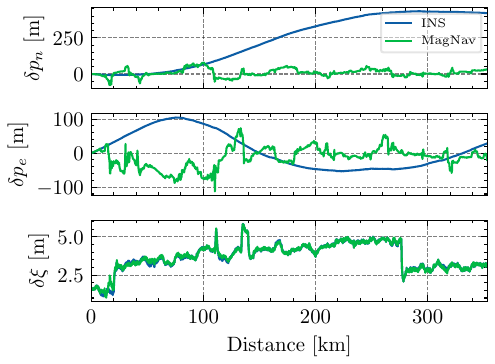}
        \caption{}
        \label{fig:pos_errors_cold_mag3}
    \end{subfigure}
    \hfill
    \begin{subfigure}[t]{0.48\linewidth}
        \centering
        \includegraphics[width=\linewidth]{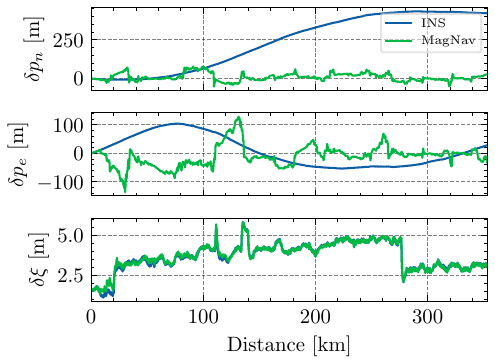}
        \caption{}
        \label{fig:pos_errors_warm_mag3}
    \end{subfigure}

    \caption{Comparison of INS and MagNav position errors (north, east, and down) for Magnetometer 3 in (a) our hybrid \ac{TL}+\ac{NN} cold-start and (b) the \ac{TL}-only cold-start scenarios.}
    \label{fig:pos_errors_cold_warm_mag3}
\end{figure}

Here we show our MagNav results along flight line 1007.06 using Magnetometer 3. \cref{fig:cold_warm_mag3_comparison} shows the evolution of the position errors in the cold- and warm start scenarios. \cref{fig:tl_parameters_mag3} shows the estimated parameters of the Tolles-Lawson and \ac{NN} model over time. \cref{fig:tl_vs_nn_output_mag3} shows their output for the correction of the magnetic field. \cref{fig:pos_errors_cold_warm_mag3} is a detailed plot of the component-wise position errors.

\clearpage
\subsection{Magnetometer 4}
\label{ssec:mag4_suppl}
\begin{figure}[tbh!]
    \centering
    \begin{subfigure}[t]{0.48\linewidth}
        \centering
        \includegraphics[width=\linewidth]{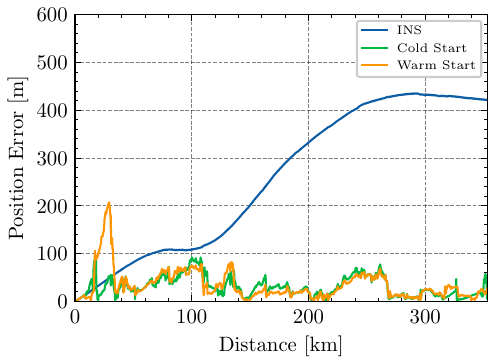}
        \caption{}
        \label{fig:cold_warm_nn_tl_mag4}
    \end{subfigure}
    \hfill
    \begin{subfigure}[t]{0.48\linewidth}
        \centering
        \includegraphics[width=\linewidth]{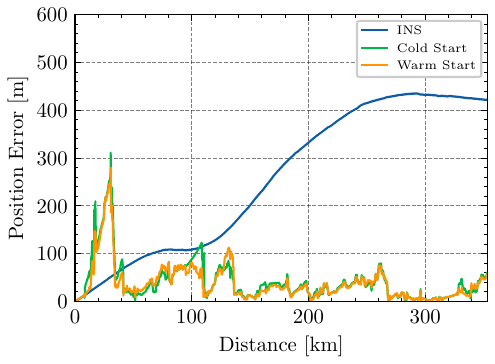}
        \caption{}
        \label{fig:cold_warm_tl_mag4}
    \end{subfigure}

    \caption{Evolution of the positioning accuracy in cold and warm start scenarios using Magnetometer 4 ($N_h = 5$). (a) Hybrid approach combining \ac{TL} and \ac{NN} calibration (\ac{DRMS} cold-start \SI{37}{\meter} / warm-start \SI{46}{\meter}) (b) \ac{TL} online calibration (\ac{DRMS} cold-start \SI{58}{\meter} / warm-start \SI{54}{\meter}).}
    \label{fig:cold_warm_mag4_comparison}
\end{figure}

\begin{figure}[tbh!]
    \centering
    \begin{subfigure}[t]{0.48\linewidth}
        \centering
        \includegraphics[width=\linewidth]{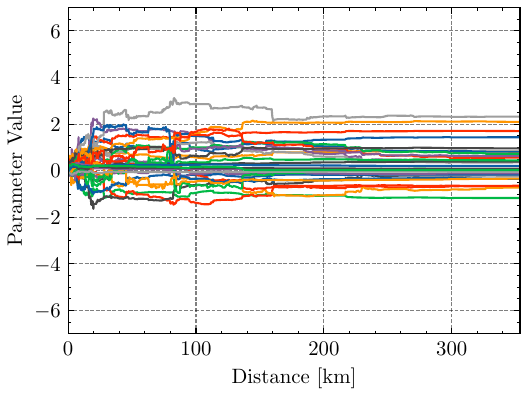}
        \caption{}
        \label{fig:nn_cold_start_mag4}
    \end{subfigure}
    \hfill
    \begin{subfigure}[t]{0.48\linewidth}
        \centering
        \includegraphics[width=\linewidth]{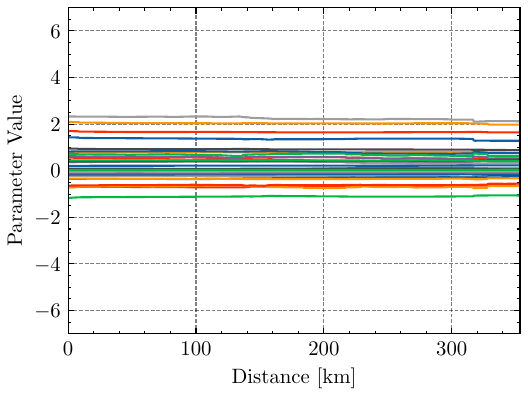}
        \caption{}
        \label{fig:nn_warm_start_mag4}
    \end{subfigure}
    \hfill
    \begin{subfigure}[t]{0.48\linewidth}
        \centering
        \includegraphics[width=\linewidth]{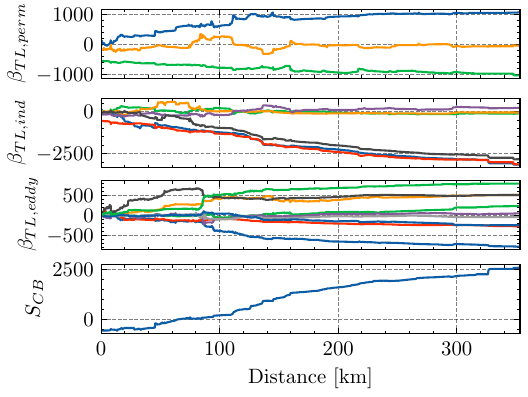}
        \caption{}
        \label{fig:tl_parameters_cold_mag4}
    \end{subfigure}
    \hfill
    \begin{subfigure}[t]{0.48\linewidth}
        \centering
        \includegraphics[width=\linewidth]{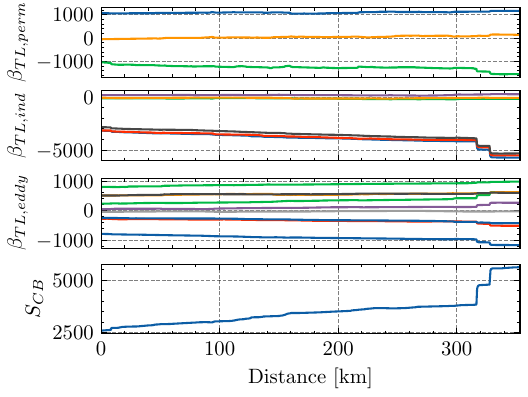}
        \caption{}
        \label{fig:tl_parameters_warm_mag4}
    \end{subfigure}

    \caption{Evolution of \ac{NN} (top) and \ac{TL} model (bottom) parameters using Magnetometer 4 ($N_h = 5$), comparing (a), (c) cold start and (b), (d) warm start.}
    \label{fig:tl_parameters_mag4}
\end{figure}

\begin{figure}[tbh!]
    \centering
    \begin{subfigure}[t]{0.48\linewidth}
        \centering
        \includegraphics[width=\linewidth]{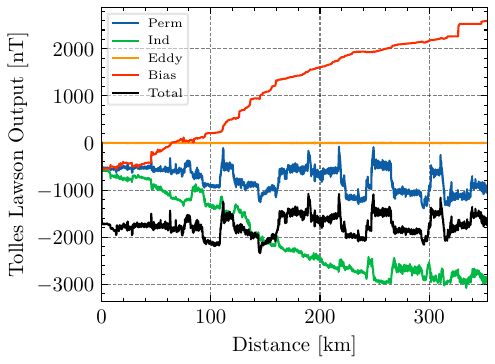}
        \caption{}
        \label{fig:tl_components_mag4}
    \end{subfigure}
    \hfill
    \begin{subfigure}[t]{0.48\linewidth}
        \centering
        \includegraphics[width=\linewidth]{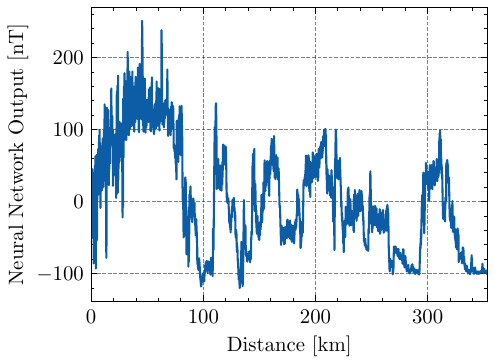}
        \caption{}
        \label{fig:nn_output_mag4}
    \end{subfigure}

    \caption{Comparison between the (a) \ac{TL} model compensation components (including the bias state $S_{CB}$) and the (b) learned \ac{NN} output using Magnetometer 4 in the cold start scenario.}
    \label{fig:tl_vs_nn_output_mag4}
\end{figure}

\begin{figure}[tbh!]
    \centering
    \begin{subfigure}[t]{0.48\linewidth}
        \centering
        \includegraphics[width=\linewidth]{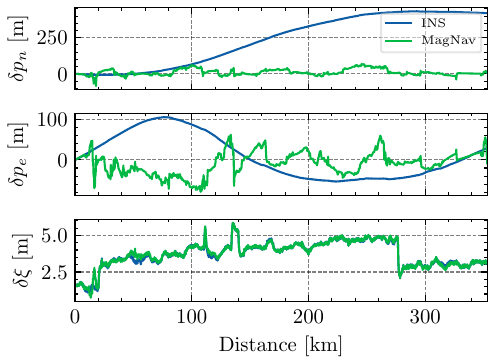}
        \caption{}
        \label{fig:pos_errors_cold_mag4}
    \end{subfigure}
    \hfill
    \begin{subfigure}[t]{0.48\linewidth}
        \centering
        \includegraphics[width=\linewidth]{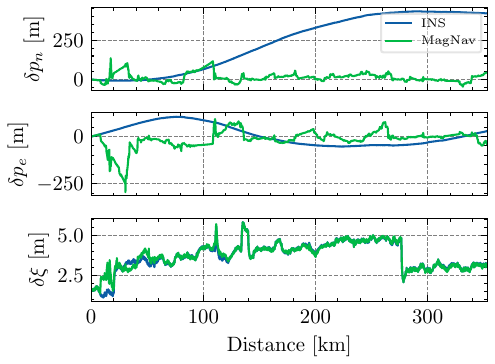}
        \caption{}
        \label{fig:pos_errors_warm_mag4}
    \end{subfigure}

    \caption{Comparison of INS and MagNav position errors (north, east, and down) for Magnetometer 4 in (a) our hybrid \ac{TL}+\ac{NN} cold-start and (b) the \ac{TL}-only cold-start scenarios.}
    \label{fig:pos_errors_cold_warm_mag4}
\end{figure}

We show the results of our MagNav algorithm for flight line 1007.06 using Magnetometer 4. \cref{fig:cold_warm_mag4_comparison} shows the evolution of the position errors in the cold- and warm start scenarios. \cref{fig:tl_parameters_mag4} shows the estimated parameters of the Tolles-Lawson and \ac{NN} model over time. \cref{fig:tl_vs_nn_output_mag4} shows their output for the correction of the magnetic field. \cref{fig:pos_errors_cold_warm_mag4} is a detailed plot of the component-wise position errors.

\clearpage
%
%
%
\subsection{Magnetometer 5}
\label{ssec:mag5_suppl}
\begin{figure}[tbh!]
    \centering
    \begin{subfigure}[t]{0.48\linewidth}
        \centering
        \includegraphics[width=\linewidth]{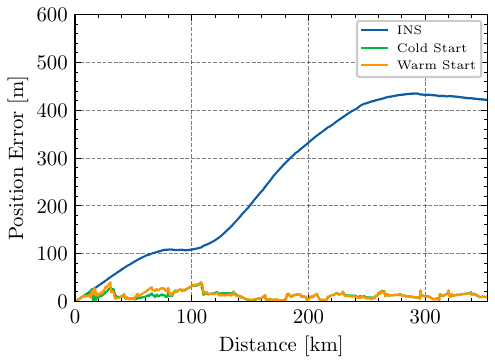}
        \caption{}
        \label{fig:cold_warm_nn_tl_mag5}
    \end{subfigure}
    \hfill
    \begin{subfigure}[t]{0.48\linewidth}
        \centering
        \includegraphics[width=\linewidth]{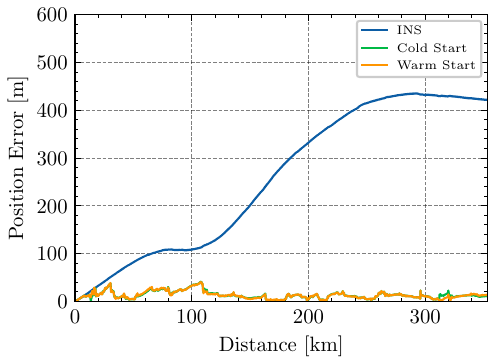}
        \caption{}
        \label{fig:cold_warm_tl_mag5}
    \end{subfigure}

    \caption{Evolution of the positioning accuracy in cold and warm start scenarios using Magnetometer 5 ($N_h = 5$). (a) Hybrid approach combining \ac{TL} and \ac{NN} calibration (\ac{DRMS} cold-start \SI{14}{\meter} / warm-start \SI{15}{\meter}) (b) \ac{TL} online calibration (\ac{DRMS} cold-start \SI{15}{\meter} / warm-start \SI{15}{\meter}).}
    \label{fig:cold_warm_mag5_comparison}
\end{figure}

\begin{figure}[tbh!]
    \centering
    \begin{subfigure}[t]{0.48\linewidth}
        \centering
        \includegraphics[width=\linewidth]{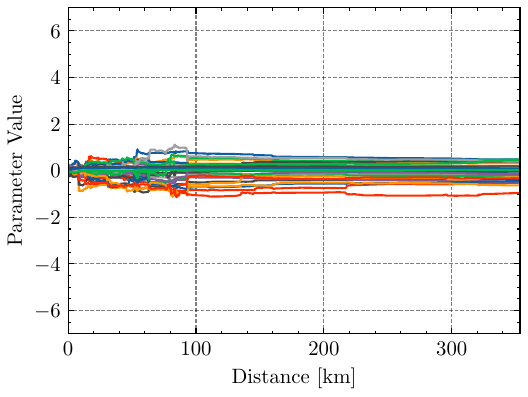}
        \caption{}
        \label{fig:nn_cold_start_mag5}
    \end{subfigure}
    \hfill
    \begin{subfigure}[t]{0.48\linewidth}
        \centering
        \includegraphics[width=\linewidth]{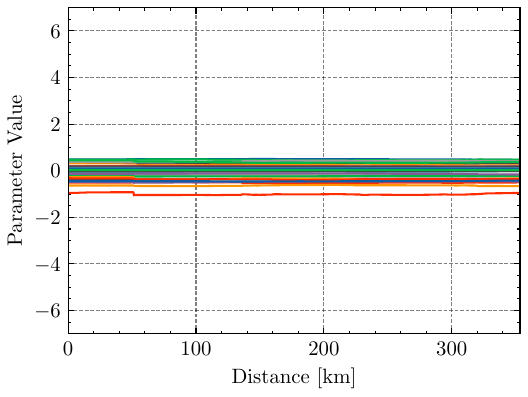}
        \caption{}
        \label{fig:nn_warm_start_mag5}
    \end{subfigure}
    \hfill
    \begin{subfigure}[t]{0.48\linewidth}
        \centering
        \includegraphics[width=\linewidth]{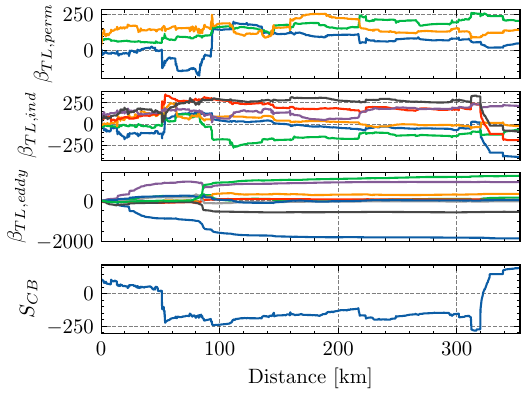}
        \caption{}
        \label{fig:tl_parameters_cold_mag5}
    \end{subfigure}
    \hfill
    \begin{subfigure}[t]{0.48\linewidth}
        \centering
        \includegraphics[width=\linewidth]{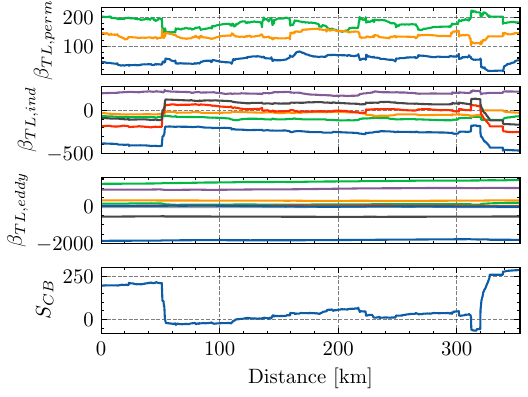}
        \caption{}
        \label{fig:tl_parameters_warm_mag5}
    \end{subfigure}

    \caption{Evolution of \ac{NN} (top) and \ac{TL} model (bottom) parameters using Magnetometer 5 ($N_h = 5$), comparing (a), (c) cold start and (b), (d) warm start.}
    \label{fig:tl_parameters_mag5}
\end{figure}

\begin{figure}[tbh!]
    \centering
    \begin{subfigure}[t]{0.48\linewidth}
        \centering
        \includegraphics[width=\linewidth]{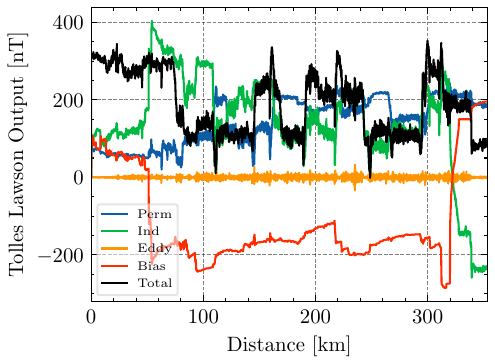}
        \caption{}
        \label{fig:tl_components_mag5}
    \end{subfigure}
    \hfill
    \begin{subfigure}[t]{0.48\linewidth}
        \centering
        \includegraphics[width=\linewidth]{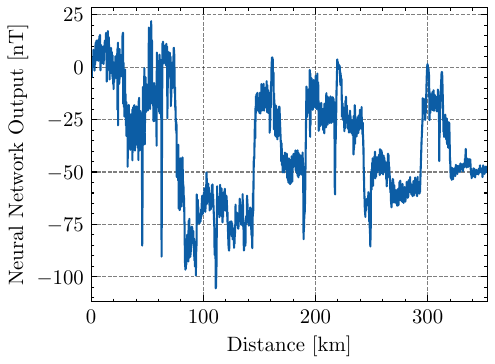}
        \caption{}
        \label{fig:nn_output_mag5}
    \end{subfigure}

    \caption{Comparison between the (a) \ac{TL} model compensation components (including the bias state $S_{CB}$) and the (b) learned \ac{NN} output using Magnetometer 5 in the cold start scenario.}
    \label{fig:tl_vs_nn_output_mag5}
\end{figure}

\begin{figure}[tbh!]
    \centering
    \begin{subfigure}[t]{0.48\linewidth}
        \centering
        \includegraphics[width=\linewidth]{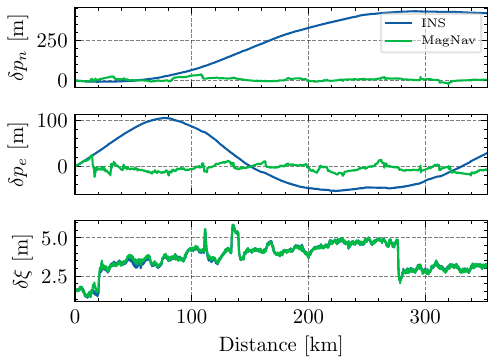}
        \caption{}
        \label{fig:pos_errors_cold_mag5}
    \end{subfigure}
    \hfill
    \begin{subfigure}[t]{0.48\linewidth}
        \centering
        \includegraphics[width=\linewidth]{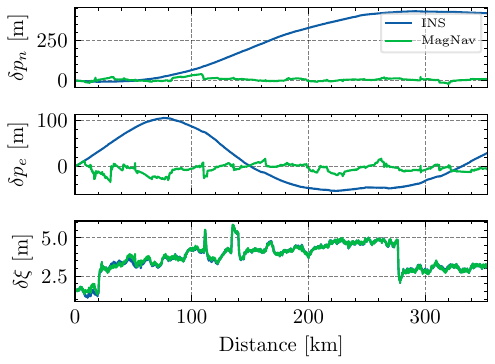}
        \caption{}
        \label{fig:pos_errors_warm_mag5}
    \end{subfigure}

    \caption{Comparison of INS and MagNav position errors (north, east, and down) for Magnetometer 5 in (a) our hybrid \ac{TL}+\ac{NN} cold-start and (b) the \ac{TL}-only cold-start scenarios.}
    \label{fig:pos_errors_cold_warm_mag5}
\end{figure}

We show the results of our MagNav algorithm for flight line 1007.06 using the low-noise Magnetometer 5. \cref{fig:cold_warm_mag5_comparison} shows the evolution of the position errors in the cold- and warm start scenarios. \cref{fig:tl_parameters_mag5} shows the estimated parameters of the Tolles-Lawson and \ac{NN} model over time. \cref{fig:tl_vs_nn_output_mag5} shows their output for the correction of the magnetic field. \cref{fig:pos_errors_cold_warm_mag5} is a detailed plot of the component-wise position errors.

\clearpage
\subsection{Remark on innovation spikes and NN activation}
We observe intermittent innovation spikes both in cold- and warm-start runs, suggesting that they are not solely initialization artifacts. After the first \SI{10}{\minute} of flight, these excursions are rejected by the $\chi^2$ innovation-based outlier rejection and thus do not enter the EKF measurement update. One hypothesis is that the current $\tanh$-based NN may suffer from saturation during online adaptation, limiting its ability to learn sharp, high-amplitude corrections from the available features. In future experiments it could be advantageous to use a \ac{NN} with non-saturating hidden-layer activations (e.g., Mish/SiLU/GELU).
\begin{figure}[tbh!]
    \centering
    \begin{subfigure}[t]{0.48\linewidth}
        \centering
        \includegraphics[width=\linewidth]{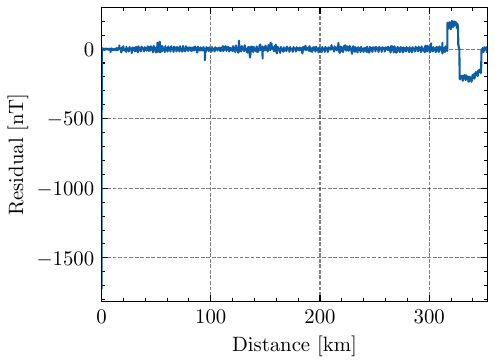}
        \caption{}
        \label{fig:residual_mag4_cold}
    \end{subfigure}
    \hfill
    \begin{subfigure}[t]{0.48\linewidth}
        \centering
        \includegraphics[width=\linewidth]{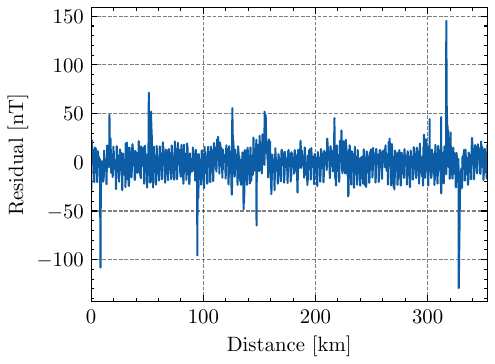}
        \caption{}
        \label{fig:residual_mag4_warm}
    \end{subfigure}

    \begin{subfigure}[t]{0.48\linewidth}
        \centering
        \includegraphics[width=\linewidth]{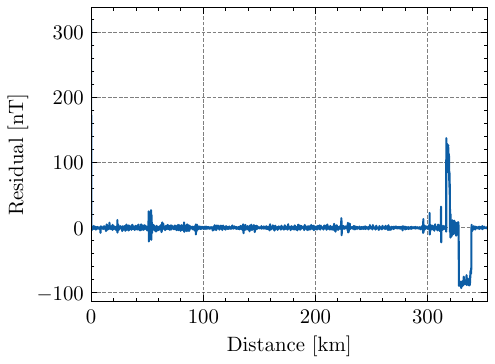}
        \caption{}
        \label{fig:residual_mag5_cold}
    \end{subfigure}
    \hfill
    \begin{subfigure}[t]{0.48\linewidth}
        \centering
        \includegraphics[width=\linewidth]{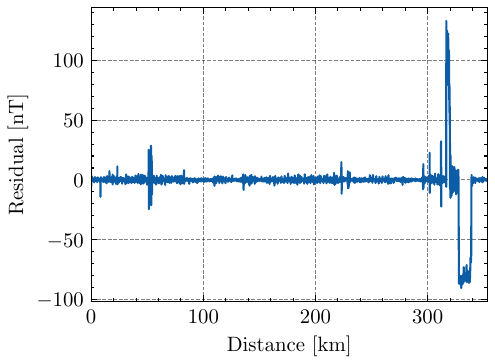}
        \caption{}
        \label{fig:residual_mag5_warm}
    \end{subfigure}

   \caption{Evolution of the innovation using ($N_h = 5$), comparing (a) Magnetometer 4 (C), (b) Magnetometer 4 (W), (c) Magnetometer 5 (C), and (d) Magnetometer 5 (W).}
   \label{fig:residual_spikes}
\end{figure}
\clearpage
\subsection{Results for Flight 1003}
\label{ssec:flight1003_results}
\begin{figure}[tbh!]
        \centering
        \includegraphics[width=\linewidth]{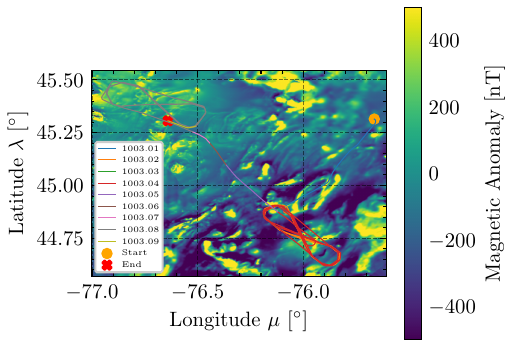}
        \caption{GNSS trajectory of all segments of flight line 1003.01 to 1003.09.}
        \label{fig:1003_plot_map}
\end{figure}

Additionally to the results for flight line 1007.06 in the main paper, we report the \ac{MagNav} positioning results for an additional trajectory from the dataset \cite{gnadt_daf-mit_2023}, flight line 1003.01 to 1003.09 with a duration of \SI{4.15}{\hour}, here. \cref{fig:1003_plot_map} shows the trajectory.

\begin{figure}[tbh!]
  \centering
    \begin{subfigure}[t]{0.48\linewidth}
        \centering
        \includegraphics[width=\linewidth]{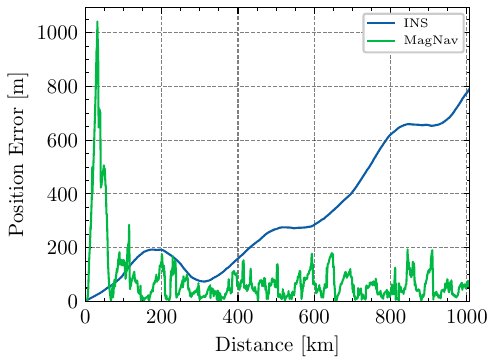}
        \caption{}
        \label{fig:1003_plot_position_error_cold}
    \end{subfigure}
    \hfill
    \begin{subfigure}[t]{0.48\linewidth}
        \centering
        \includegraphics[width=\linewidth]{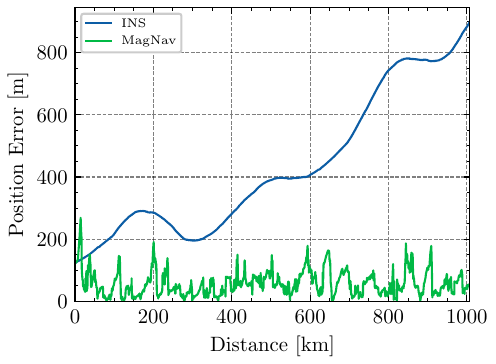}
        \caption{}
        \label{fig:1003_plot_position_error_warm}
    \end{subfigure}

  \caption{Evolution of the positioning accuracy for our hybrid approach combining \ac{TL} and \ac{NN} calibration using Magnetometer 3 ($N_h = 2$) in the (a) cold-start (\ac{DRMS} \SI{142}{\meter}) and (b) warm-start (\ac{DRMS} \SI{70}{\meter}) scenarios.}
  \label{fig:1003_plot}
\end{figure}

In \cref{fig:1003_plot} we show the time evolution of the position error of our method in the cold ( \cref{fig:1003_plot_position_error_cold}) and warm (\cref{fig:1003_plot_position_error_warm}) start scenarios.

The cold-start is initialized as described in the main publication. The warm-start was initialized with an error of ca. \SI{122}{\meter}, since the INS in the dataset drifted some time before the magnetometer data was available.
Magnetometer 2 exhibited periodic signal dropouts during pitch/roll maneuvers, particularly affecting line 1003.01. Therefore, Magnetometer 3---though less noisy, but without dropouts---is used here. 
For areas not covered by the magnetic anomaly maps in \cite{Gnadt2023, gnadt_daf-mit_2023}, the \emph{Canada -- 200\,m Residual Magnetic Field -- 2025 -- Aug} dataset \cite{GSC_2025_ResidualMag_200m} was used. 